\documentclass[a4paper,fleqn]{cas-sc}
\usepackage{booktabs}  
\usepackage{array}     
\usepackage{tabularx}  
\usepackage{geometry} 
\usepackage{longtable}
\usepackage{graphicx}
\usepackage{subcaption}
\usepackage{float}
\usepackage{tabularray}
\usepackage[square,numbers,super,sort&compress]{natbib}
\usepackage{adjustbox}
\def\tsc#1{\csdef{#1}{\textsc{\lowercase{#1}}\xspace}}

\tsc{WGM}
\tsc{QE}
\tsc{EP}
\tsc{PMS}
\tsc{BEC}
\tsc{DE}

\begin{document}
\let\WriteBookmarks\relax
\def\floatpagepagefraction{1}
\def\textpagefraction{.001}
\shorttitle{}
\shortauthors{J.Li et~al.}

\title [mode = title] {An Electrocardiogram Foundation Model Built on over 10 Million Recordings with External Evaluation across Multiple Domains}                      

\tnotetext[1]{Work on this project was done by the corresponding author (Shenda Hong) was a visiting scholar at Massachusetts General Hospital (MGH) and during a research internship by the first author (Jun Li) at Beth Israel Deaconess Medical Center (BIDMC).}

\author[1,9,10]{Jun Li}
\author[2,5]{Aaron Aguirre}
\author[2,5]{Junior Moura}
\author[7]{Che Liu}
\author[8]{Lanhai Zhong}
\author[5,6]{Chenxi Sun}
\author[3,4]{Gari Clifford}
\cormark[1]
\author[5,6]{Brandon Westover}
\cormark[1]
\author[1,9,10]{Shenda Hong}
\cormark[2]



\affiliation[1]{organization={National Institute of Health Data Science, Peking University},
                state={Beijing},
                country={China}}

\affiliation[2]{organization={Department of Cardiology, Massachusetts General Hospital},
                state={Boston, MA},
                country={USA}}

\affiliation[3]{organization={Department of Biomedical Informatics, School of Medicine, Emory University},
                city={Atlanta},
                state={GA}, 
                country={USA}}

\affiliation[4]{organization={Department of Biomedical Engineering, Georgia Institute of Technology},
                city={Atlanta},
                state={GA}, 
                country={USA}}

\affiliation[5]{organization={Harvard Medical School},
                city={Boston},
                state={MA}, 
                country={USA}}

\affiliation[6]{organization={Department of Neurology, Beth Israel Deaconess Medical Center},
                city={Boston},
                state={MA}, 
                country={USA}}

\affiliation[7]{organization={Department of Computing, Data Science Institute, Imperial College London},
                city={London},
                country={UK}}

\affiliation[8]{organization={Zhongshan School of Medicine, Sun Yat-sen University},
                city={Guangzhou},
                country={China}}                

\affiliation[9]{organization={Institute of Medical Technology, Peking University Health Science Center},
                city={Beijing},
                country={China}}                

\affiliation[10]{organization={Institute for Artificial Intelligence, Peking University},
                city={Beijing},
                country={China}}

\cortext[cor1]{co-senior authors}
\cortext[cor2]{Corresponding author: Shenda Hong, hongshenda@pku.edu.cn}

\begin{abstract}
\textbf{BACKGROUND}. Artificial intelligence (AI) has demonstrated significant potential in ECG analysis and cardiovascular disease assessment. Recently, foundation models have played a remarkable role in advancing medical AI, bringing benefits such as efficient disease diagnosis and cross-domain knowledge transfer. The development of an ECG foundation model holds the promise of elevating AI-ECG research to new heights. However, building such a model faces several challenges, including insufficient database sample sizes and inadequate generalization across multiple domains. Additionally, there is a notable performance gap between single-lead and multi-lead ECG analyses.

\textbf{METHODS}. We introduced an ECG Foundation Model (ECGFounder), a general-purpose model that leverages real-world ECG annotations from cardiology experts to broaden the diagnostic capabilities of ECG analysis. ECGFounder was trained on over 10 million ECGs with 150 label categories from the Harvard-Emory ECG Database, enabling comprehensive cardiovascular disease diagnosis through ECG analysis. The model is designed to be both an effective out-of-the-box solution, and a to be fine-tunable for downstream tasks, maximizing usability. Importantly, we extended its application to lower rank ECGs, and arbitrary single-lead ECGs in particular. ECGFounder is therefore applicable to supporting various downstream tasks in mobile and remote monitoring scenarios.

\textbf{RESULTS}. Experimental results demonstrate that ECGFounder achieves expert-level performance on internal validation sets, with AUROC exceeding 0.95 for eighty diagnoses. It also shows strong classification performance and generalization across various diagnoses on external validation sets. When fine-tuned, ECGFounder outperforms baseline models in demographic analysis, clinical event detection, and cross-modality cardiac rhythm diagnosis, surpassing baseline methods by 3 to 5 points in AUROC.

\textbf{CONCLUSIONS}. The ECG foundation model offers an effective solution, allowing it to generalize across a wide range of tasks. By enhancing existing cardiovascular diagnostics and facilitating integration with cloud-based systems that analyze ECG data uploaded from wearable devices, it significantly contributes to the advancement of cardiovascular AI community and enables management of cardiac conditions. Code and model can be found at \url{https://github.com/PKUDigitalHealth/ECGFounder}.

\end{abstract}


\maketitle

\section{Introduction}

The Electrocardiogram (ECG) is a crucial diagnostic tool that measures and records the electrical activity of the heart using electrodes placed on the skin.\cite{siontis2021artificial} ECG recordings are essential for diagnosis and monitoring of cardiac health conditions. However, fully interpreting an ECG is complex and requires significant training and time. A typical ECG expert undergoes nearly 10 years of training, including medical school, internal medicine residency, and specialized ECG training.\cite{berkaya2018survey} 
In recent years, the advent of deep learning together with efforts to assemble relatively large databases of ECGs have seen some interesting progress in the field, extending ECG analysis beyond the traditional medical domains \cite{Attia2019,attia2019artificial,zhu2024four,zhuAutomaticMultilabelElectrocardiogram2020}. 
However, due to the lack of large-scale publicly available ECG databases with diverse diagnostic information, developing a general-purpose AI-ECG model remains a challenging task. Existing models are often confined to specific diagnostic tasks and datasets. This challenge highlights the practical limitations of training an ECG model from scratch on small-scale datasets, making it difficult to extend the model to real-world ECG analysis. 

Foundation models, with their strong generalization capabilities, provide a promising approach for enhancing the performance of AI-ECG models through transfer learning. Recently, these models have achieved significant advancements in the field of medical AI. In retinal disease diagnosis, the RETFound model, through pre-training on a large number of retinal images, has achieved excellent performance across various clinical diagnostic tasks.\cite{zhouFoundationModelGeneralizable2023a} In computational pathology, the UNI model was trained on a vast amount of whole-slide images, reaching expert-level performance in multiple cancer diagnostic tasks.\cite{chenGeneralpurposeFoundationModel2024}  In these studies, foundation models are defined as large-scale AI models trained on extensive datasets, capable of adapting to a wide range of downstream tasks. Specifically, they meet the following criteria: 1) the pre-training dataset is large in scale, 2) the model has an enormous number of parameters, and 3) it can perform a wide range of downstream tasks.\cite{bommasani2021opportunities}

There have been several claims in the literature of developing foundation models for the ECG.\cite{yu2024ecg,mckeen2024ecg}
However, due to limitations of existing ECG databases in terms of sample size, patient numbers, the variety of diagnoses, and importantly, the demographics of the patients, these models have yet to address the challenges of diversity across regions, ethnic groups, and diagnostic variations.\cite{MERDJANOVSKA2022117206} 
Moreover, to qualify as a foundation model, the trained model must be capable of generalizing to multiple domains outside of the initial training paradigm. \cite{GariClifford_2024_10_11_past_present}
Additionally, current ECG models have significant performance degradation in single-lead ECGs compared to multi-lead ECGs.\cite{MatthewAReyna_2021_09_13_Willtwo}


In this work, we propose the ECG Foundation Model, ECGFounder, which is capable of diagnosing 150 types of cardiac abnormalities, leveraging over ten million clinically annotated real-world ECGs spanning all existing ECG classifications. This represents the largest and most comprehensive ECG foundation model to date. Moreover, we demonstrate how the model is applicable to a wide range of tasks in varying domains, a fundamental requirement for foundation models.
We provide the medical machine learning community with an accessible ECG foundation model that offers an effective out-of-the-box solution and fine-tuning capabilities. Compared to traditional AI-ECG models trained from scratch, ECGFounder offers a novel approach that achieves superior performance through fine-tuning. This advancement has the potential to drive the future development of AI-ECG technology. 

To address the inherent challenges of incomplete annotations in real-world data, we introduce a novel method for pre-processing and training on these annotations, ensuring robust performance even under sub-optimal conditions. Moreover, by training the single-lead ECG model based on lead augmentation, we are able to maintain high diagnostic performance on single-lead ECGs as well. We validate the model's performance on both internal and external test sets, where it consistently matches expert-level diagnoses. In downstream task fine-tuning, we demonstrate ECGFounder's versatility in addressing various tasks, including demographics detection, clinical event detection, and cross-modality diagnosis (Figure 1b). Specifically, we evaluate ECGFounder on 12 clinical tasks, such as ECG age regression and classification, sex detection, chronic kidney disease (CKD) detection, chronic heart disease (CHD) detection, regression and abnormal classification of left ventricular ejection fraction (LVEF), regression and abnormal classification of NT-proBNP, and atrial fibrillation detection based on photoplethysmography (PPG). The results of these downstream tasks highlight ECGFounder’s potential as a foundational model for the further development of AI-ECG models. 

\begin{figure*}[]
    \centering
    \includegraphics[width=0.9 \textwidth]{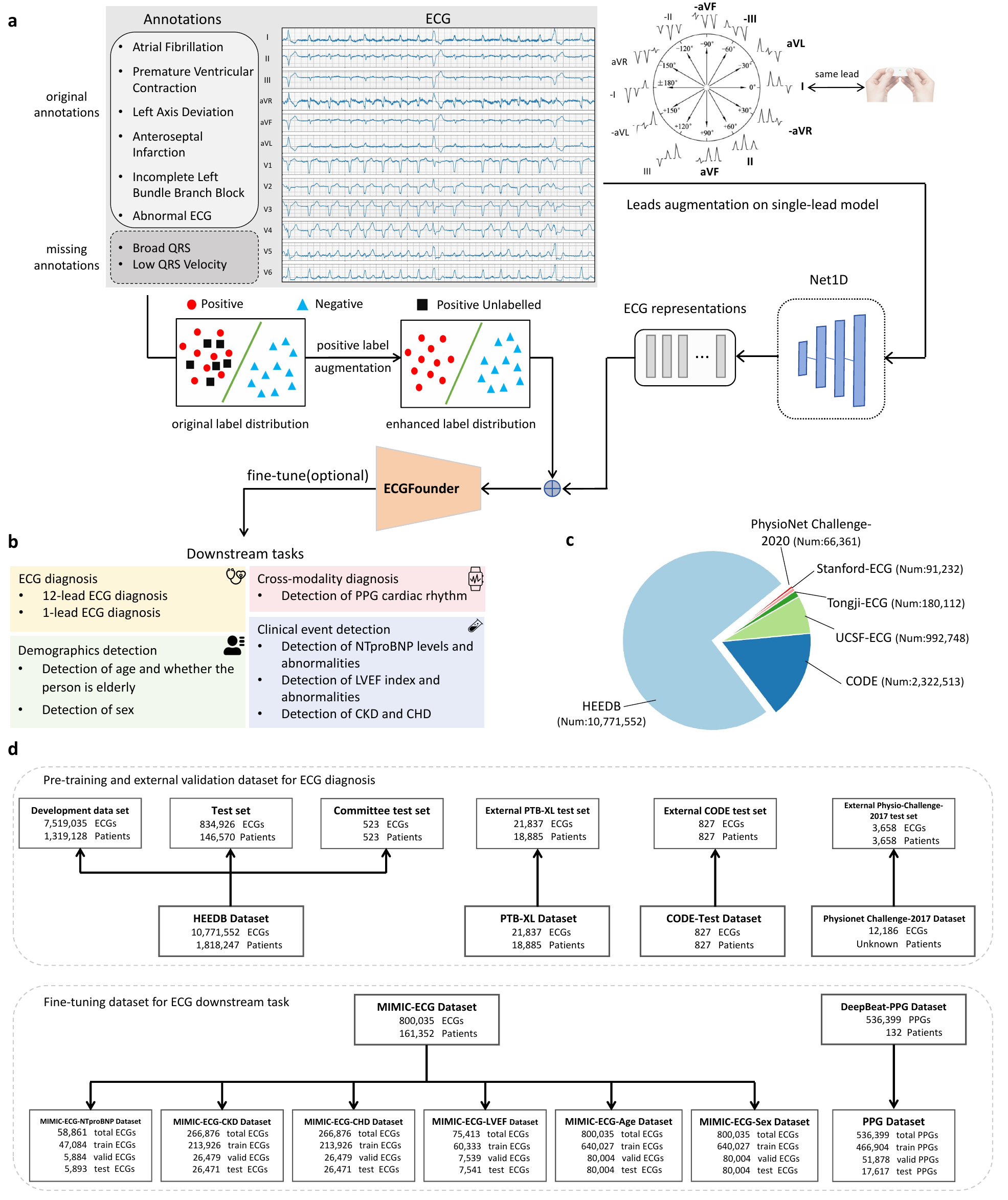} 
    \caption{ECGFounder is a general ECG encoder based on the RegNet architecture. (a) ECGFounder was trained on 10,771,552 ECGs with 150 types of ECG diagnostic labels. Due to the presence of missing diagnoses in real-world expert annotations, we implemented positive label augmentation by modifying the loss function of the pre-training method. For more details, see the Methods section. (b) ECGFounder was applied to 12 clinical downstream tasks, covering ECG diagnosis, demographics inference, clinical event detection, and cross-modality diagnosis. In comparison with baseline methods, ECGFounder achieved state-of-the-art performance across all tasks. (c) A comparison of the HEEDB dataset used by ECGFounder with other large ECG datasets. (d) Data used in model development and validation for the ECG diagnosis and downstream tasks. PPG., photoplethysmography; LVEF., left ventricular ejection fraction; CKD., chronic kidney disease; CHD., chronic heart disease.}
    \label{fig:framework}
\end{figure*}

\section{Methods}

\subsection{Datasets and pre-processing}

Our dataset, the Harvard-Emory ECG Database (HEEDB), is currently the largest open-access ECG dataset, containing 10,771,552 expert-annotated ECGs from 1,818,247 unique subjects.\cite{koscovaHarvardEmoryECGDatabase2024} These ECGs are predominantly 10-second, 12-lead clinical ECGs. 
The dataset includes annotations from cardiologists and ECG technicians paired with the ECGs. These annotations consist of discrete text reports that primarily describe the morphology, rhythm, and diagnostic information of the ECGs. Experts used the  Marquette 12SL ECG Analysis Program (GE Healthcare) version 4 to assist with annotations, which provides waveform parameters for doctors' reference.\cite{healthcare2007marquettetm} The program offers many diagnostic categories, allowing doctors to simply click on corresponding category labels on the computer, avoiding manual entry of diagnostic statements. To extract descriptive information about ECGs from these discrete labels, we utilized regular expressions to parse the annotations, tallying each independent label. In total, there were 287 independent phrases. After reviewing with doctors, we removed phrases that were irrelevant to ECG descriptions while retaining meaningful phrases. Ultimately, we defined 150 meaningful labels, including rich diagnostic information as well as specific rhythm and morphological descriptions (See Supplement for more details about labels).

Our external validation data comprised three large ECG databases (DB): The CODE-test DB \cite{ribeiroAutomaticDiagnosis12lead2020}, the PTB-XL DB \cite{wagnerPTBXLLargePublicly2020}, and PhysioNet Challenge-2017 DB \cite{clifford2017af}. The CODE-test DB is composed of ECG records from 827 patients across 811 municipalities in Brazil, collected by the Telehealth Network of Minas Gerais (TNMG). There are six common arrhythmia labels in this ECG DB, annotated by several experienced ECG experts.

The PTB-XL dataset contains 21,837 clinical ECGs from 18,885 patients in Germany. Each ECG is a 10-second, 12-lead recording. The labels were reviewed and verified by two physicians. This dataset is currently one of the best publicly accessible ECG collections, both in terms of the number of samples and the quality of the labels.\cite{wagnerPTBXLLargePublicly2020}

The PhysioNet Challenge-2017 is a large single-lead ECG dataset. The ECG recordings were collected using the AliveCor device.\cite{clifford2017af} The training set includes 8,528 single-lead ECG recordings, with durations ranging from 9 to 60 seconds, while the test set contains 3,658 ECG recordings of similar lengths. The dataset requires classification of ECG recordings into normal rhythm, atrial fibrillation rhythm, other rhythms, and noise.

Additionally, we utilized the MIMIC-IV-ECG DB to fine-tune our model for various downstream tasks. The MIMIC-IV-ECG dataset is part of the MIMIC series, focusing on the collection and analysis of ECG data.\cite{gow2023mimic,physionet_doi:10.1161/01.CIR.101.23.e215} MIMIC-IV-ECG originates from real clinical settings at Beth Israel Deaconess Medical Center (BIDMC) in Boston, USA, and contains 800,035 clinical ECGs from 161,352 patients treated in the intensive care unit (ICU). Moreover, ECG recordings in the dataset can be matched with the electronic health records of patients in the MIMIC-ED, allowing ECG data to be associated with specific conditions. Here, we linked several clinical downstream tasks, such as age, sex, chronic kidney disease (CKD), chronic heart disease (CHD), left ventricular ejection fraction (LVEF) and NT-proBNP  with ECG data to explore the performance improvement of the fine-tuned ECGFounder model in detecting other diseases and clinical events. More details about the split and labels of MIMIC-IV-ECG DB can be found in Supplement. Proportions of positive and negative cases naturally reflect clinical prevalence but were not deliberately controlled or balanced.

To explore ECGFounder's generalization capabilities on other similar physiological signals, we fine-tuned and evaluated it using the DeepBeat dataset, a PPG-based atrial fibrillation (AF) detection dataset.\cite{torres2020multi} The dataset includes over 500,000 labeled signals from more than 100 individuals.

For data preprocessing, unreadable files, missing data, and unmatched data were excluded. 
Our final development dataset includes 7,519,035 ECGs from 1,319,128 patients, and the 
test dataset includes 834,926 ECGs from 146,570 patients. We applied linear interpolation 
to resample ECG frequencies to 500 Hz. We used a high-pass filter with a cutoff frequency 
of 0.5 Hz to suppress residual baseline drift and a second-order 50 Hz Butterworth low-pass 
filter to reduce high-frequency noise. A 50/60-Hz notch filter was utilized to eliminate 
electrical interference. For ECG records longer than 10 seconds, we extracted 10-second 
windows in sequence. If a sequence was less than 10 seconds, we applied zero padding. 
We normalized all signals using the mean and standard deviation of each individual signal 
segment before inputting them into the model.

\subsection{Model architecture}
To establish the model, we used an architecture tailored for ECG, capable of learning 
generalizable representations from large-scale ECG datasets. The increase of ECG data and 
the number of leads meant that the model must not only consider temporal information, but 
also spatial relationships (i.e., interactions between different leads and the overall pattern of cardiac electrical activity). This is crucial for ensuring that the ECG foundation model is applicable to real-world clinical ECG scenarios, as it mitigates the impact of nonuniform ECG durations and missing leads in the training dataset.

Considering these factors, we built our model architecture based on our previously proposed Net1D.\cite{hong2020holmes} It is built on top of the Self-Regulated Network for Image Classification (RegNet) architecture.\cite{radosavovic2020designing} This structure begins with a stage-wise network design where each stage consists of a set number of blocks and channels that scale with network depth. This is beneficial for us to expand and design the blocks and channels of the ECG foundation model. Unlike traditional uniform scaling across layers, RegNet employs a quantized linear model to predict optimal widths and depths, ensuring efficient performance across a range of model sizes. The initial layers focus on capturing low-level features with fewer channels, which gradually increase as the network deepens, thus optimizing computational efficiency and model capacity. Following this, the model utilizes a series of bottleneck blocks that combine group convolutions and channel-wise attention mechanisms, enhancing the richness of representation in both temporal and spatial dimensions while controlling model complexity. This tailored configuration makes the model suitable for ECG data. More details about the model architecture can be found in the Supplement.

\subsection{Training with real-world annotations: noisy, imbalanced, positive unlabeled}
Unlike conventional ECG diagnostic models that typically use single-label classification methods, we employed multi-label classification during the training phase of the ECG foundation model. This approach aligns more closely with clinical practice, where an ECG diagnosis often consists of multiple diagnostic labels. For example, an abnormal ECG diagnosis might be something like normal sinus rhythm | premature ventricular complexes | premature ectopic complexes. Additionally, multi-label diagnostics better meet the application needs of clinicians. During training, utilizing a multi-label classification approach provides our model with rich semantic value and facilitates generalization to different annotation systems. 

However, the nature of multi-label annotations means that it is almost impossible to achieve perfect multi-label data, especially in cases like ECG where there are many diagnostic categories. Generally speaking, if a cardiology expert can annotate three classes simultaneously, it is considered excellent. However, the actual number of ECG diagnostic labels far exceeds three classes.\cite{kashouECGInterpretationProficiency2023} 

To address the challenge brought by incomplete ECG annotations, we introduce positive unlabeled (PU) learning method. PU learning is defined as when positive samples are present in the data but unlabeled, but the labeled positive samples are correct, meaning that the unlabeled samples are not necessarily negative examples.\cite{zhao2022dist} Conventional multilabel classification methods typically treat unlabeled categories as negative by default, but this assumption does not hold true for ECG data.

Positive unlabeled learning in the ECG context leads to a severe imbalance in the predicted 
probability distribution: an ECG usually contains a few positive diagnostic labels, with 
the remaining labels being treated as negative. This creates a severe positive–negative 
imbalance, where negative samples far outnumber positive samples for each label. When 
using conventional loss functions, such as the binary cross-entropy loss function, the model 
tends to learn from simpler samples, namely the true negative samples, while the more 
challenging samples, namely the false negative samples (which may in fact be missed 
positives), are harder to fit. As a result, the model’s predicted probabilities tend to skew 
toward 0 rather than 1, diminishing its ability to detect clinically important abnormalities.

To enhance the model's ability to fit representations of positive samples, we improved the multi-label classification loss function, enabling the model to correct missing labels by adjusting the weights of positive and negative samples. Typically, for missing labels, a well-trained multi-label model's predicted probability should be close to 1 rather than 0. Therefore, by applying a smaller weight to the loss of negative labels with predicted probabilities close to 1, we can mitigate the impact of missing labels. The loss function is given by:

\begin{equation}
    \mathcal{L}=-(\gamma-p)p^{2}.
    \label{loss}
\end{equation}

Here, $\gamma$ is a hyperparameter of the model and $p$ is the predicted probability of model. In this case, it is set to 
$\gamma = 1.5$, which we find optimally balances the weights, allowing the model to learn good representations of both positive and negative samples.

Our model training used AdamW to minimize the loss function, with an initial learning rate set to 0.001. The learning rate decayed by a factor of 10 every 5 epochs. The trainable temperature parameter was initialized to 0. Training spanned a maximum of 20 epochs, with early stopping based on validation loss. We used a batch size of 1024.

\subsection{Training a single-lead ECG model based on leads augmentation}
In recent years, portable and wearable ECG-device development has revolutionized 
continuous cardiac monitoring, offering a noninvasive method for real-time assessment. 
Beyond the conventional diagnosis of cardiac arrhythmias, another critical challenge in this 
field is accurately detecting and interpreting ECG axis deviation on single-lead ECGs from 
wearable devices (typically lead I), which can significantly impact the diagnosis of various 
cardiac abnormalities. For instance, left axis deviation can provide additional insights 
into diagnosis, such as left ventricular hypertrophy, left bundle-branch block, left anterior 
fascicular block, preexcitation syndromes, and inferior myocardial infarction (MI).\cite{kamga2022use} 

To address this issue, we developed a novel training method utilizing lead-augmented wearable ECG models. By systematically enhancing standard 12-lead ECG data, we simulate various clinical scenarios of axis inversion, thereby enhancing the model's robustness and versatility. Understanding the relationship between ECG vectors and leads is crucial for this method. The cardiac electrical activity generates a vector representation of cardiac signals, captured by different ECG leads placed on the body. Each lead provides a unique perspective of the cardiac electrical axis, offering a comprehensive view when combined. The standard 12-lead ECG system includes limb leads (I, II, III, aVR, aVL, aVF) and precordial leads (V1-V6), with each lead representing a specific projection of the cardiac electrical vector.\cite{meek2002abc} Specifically, we primarily utilize the ECG signals from limb leads. Based on the angular position of each limb lead's axis relative to the heart, we consider lead I as the center of the semicircle (i.e., 0°) and calculate the signals from all leads around the semicircle (i.e., from -90° to 90°), thereby obtaining six additional leads (aVL, -aVR, II, -III, aVF, -aVF). We then trained a model for wearable ECG devices, extracting the lead I ECG from the HEEDB 12-lead data and randomly incorporating one of the remaining six augmented leads into the model for training with a 50\% probability. This ensures that the model can learn arrhythmia features from lead I data and axis deviation from the additional six leads. Additionally, we have scaled down the model's parameter size to optimize for wearable devices with limited computational resources.

\subsection{Fine-tuning on ECG foundation model}
When adapting to specific ECG downstream tasks, we needed to retain the parameters of 
the base model and discard the initial classification linear layer. The number of classes in 
the downstream task determines the number of neurons needed in the final layer of the new 
linear layer. The training objective is to generate classification outputs that match the labels. We adopted two different methods of fine-tuning: linear probing and full fine-tuning. During the linear probing experiments, we only fine-tuned the parameters of the linear classification head on top of the pretrained model, keeping all other pretrained model weights frozen. During full fine-tuning, we allowed all pretrained model weights to be updated and adapted to the downstream classification tasks.

The total training period is 30 epochs, with a learning rate adjustment strategy that utilizes the scheduler. After every epoch, the scheduler monitors the specified metric, and if the performance does not improve for 10 consecutive epochs, the learning rate is reduced by a factor of 0.1. The learning rate reduction is triggered based on the maximization of the monitored metric. This approach ensures a dynamic adjustment of the learning rate depending on the training progress. After training in each epoch, the model is evaluated on the validation set.\cite{chenGeneralpurposeFoundationModel2024} The model weights with the highest AUC on the validation set are saved as a checkpoint for future evaluation.

\subsection{Clinical validation}
To validate and compare the performance of our model, we followed the committee 
experimental design of Hannun et al.\cite{hannunCardiologistlevelArrhythmiaDetection2019} We established the committee consisting of three experienced ECG cardiologists to annotate a subset of the internal test set, which includes the 523 most recent ECGs from 523 unique patients. We developed an ECG annotation system for cardiologists and 20 label types, including cardiac-rate abnormalities, conduction blocks, myocardial dilation, MI, and ECG morphologies, with sublabels under each category. Table 1 displays the complete list of label types. After independent annotation by the committee, a consensus determination was made; labels that did not reach consensus were removed, providing an expert standard for model evaluation. Labels that the committee could not interpret or agree on were eliminated from our test dataset.

In addition, to compare the diagnostic accuracy between the model and cardiologists, four 
additional ECG cardiologists were involved and provided specific instructions on how to 
use the system. Table S6 shows the cardiologists’ ages, levels of experience, and education. 
Each cardiologist was required to annotate each ECG from the previous internal test set. The 
annotations from these cardiologists were then compared with the model’s results.

The evaluation of the model was conducted by calculating accuracy, the area under the Receiver Operating Characteristic curve (AUROC), sensitivity, and specificity. 

\section{Results}
Experimental results demonstrate that ECGFounder achieves superior performance on 
internal validation sets of 12-lead ECGs, with AUROC exceeding 0.95 for 80 diagnoses 
(Fig. S3). We further validated ECGFounder for 12-lead ECGs using the committee’s 
internal test set. The algorithm’s average AUROC score for diagnosing all 20 classifications 
was 0.968 (95\% confidence interval [CI], 0.955 to 0.982), sensitivity was 0.971 (95\% CI, 
0.639 to 0.988), and specificity was 0.937 (95\% CI, 0.912 to 0.953). When comparing the 
model with cardiologists, our model achieved an overall average F1 score of 0.677 (95\% CI, 
0.480 to 0.802), outperforming the cardiologists’ average F1 score of 0.640. The model’s 
performance was compared with cardiologists’ performance across 20 diagnostic categories 
(Table 1). When comparing the model’s receiver operating characteristic curve with the 
true-positive rate and false-positive rate of cardiologists, the model outperformed the average performance of cardiologists for most labels (Fig. S2).

In the external test set experiments for 12-lead ECGs, we evaluated the performance of 
our model and other models on the CODE-test and PTB-XL datasets. On CODE-test, our model achieved an average AUROC of 0.981 (95\% CI, 0.979 to 0.984), outperforming other 
baseline models such as S12L-ECG, which had an average AUROC of 0.980 (95\% CI, 0.978 
to 0.982); CTN, which had an average AUROC of 0.963 (95\% CI, 0.960 to 0.967); and 
ECG Squeeze-and-Excitation Residual Neural Network (ECG-SE-ResNet), which had an 
average AUROC of 0.963 (95\% CI, 0.961 to 0.967) (Table 2). On PTB-XL, as the other two 
models can only diagnose arrhythmias and cannot complete other diagnostic classifications, 
we validated this dataset only on the class that the models have. Our model achieved an 
average AUROC of 0.924 (95\% CI, 0.917 to 0.931) (Table 2).\cite{ribeiroAutomaticDiagnosis12lead2020,natarajan2020wide,zhu2020classification} These results indicate that our model generalizes to different regions, hospitals, and patients.

Internal test set experiments focused on single-lead ECGs, with the model demonstrating 
excellent performance in rhythm-type diagnosis (Fig. S3). It achieved an AUROC above 
0.95 for common heart-rate abnormalities such as normal sinus rhythm, sinus bradycardia, 
sinus tachycardia, marked sinus bradycardia, sinus arrhythmia, marked sinus arrhythmia, 
and AF. These diagnoses can be reliably identified from single-lead ECGs. In addition, 
the model achieved an AUROC above 0.8 for diagnosis, including premature ventricular 
complexes, premature supraventricular complexes, pacemaker, first-degree atrioventricular 
block, branch block, fascicular block, and chamber enlargement. The performance observed 
for some of these diagnoses was exceptional for the use of single-lead ECGs and is worthy 
of note. In MI diagnosis, the model also showed good diagnostic performance for lateral 
infarct, anterolateral infarct, acute MI, and ST-segment elevation MI. In ECG research, 
these diagnoses are associated with shifts in the heart’s electrical axis. This suggests that 
data-augmentation methods based on the electrical axis are effective for training single-lead 
ECG models.

In the external test set experiments on single-lead ECG devices, our model accurately 
classified sinus rhythm and AF. The model’s performance, as shown in supplement, achieved 
an AUROC of 0.975 (95\% CI, 0.972 to 0.977) and 0.957 (95\% CI, 0.955 to 0.959) for 
normal sinus rhythm and AF, respectively. These results demonstrate excellent performance 
in analyzing ECG signals collected from portable and wearable ECG devices under real-world conditions. It should be emphasized that ECGFounder is executed on a cloud-based 
system, analyzing data uploaded from wearable devices, rather than operating directly on the 
wearable devices themselves.

We next validated the performance of ECGFounder in transfer learning. Our model was 
fine-tuned and evaluated on six downstream tasks using supervised learning on the MIMIC-IV-ECG dataset, resulting in both a 12-lead ECG model and a single-lead ECG model. 
The fine-tuning results are shown in Figure 2. As shown, ECGFounder outperforms the 
baseline methods in every downstream task. Specifically, it achieves 2 to 3 percentage points 
higher performance than ECG-SimCLR and 4 to 6 percentage points higher than ECG- in 
age, sex, NT-proBNP, LVEF, CKD, and CHD detection. In addition, for comparison with 
the previously published CKD study,\cite{holmstromDeepLearningbasedElectrocardiographic2023} we created an independent test set consisting of 
individual patients from the MIMIC-IV-ECG dataset for validation. The results are shown in 
Table 3. It should be noted that ECGFounder was the internal validation, while the compared 
method was the external validation.

\begin{table*}[h]
\caption{Performance of ECGFounder and cardiologists on the committee test set}
\begin{tabular}{ll|cccc|ccc}
\toprule
\label{tab:internal}
{} & {} & \multicolumn{4}{c}{ECGFounder} & \multicolumn{3}{c}{Cardiologists} \\ \midrule
 
Class                                & Count &  AUC &  Sens &  Spec & F1 & Sens &  Spec & F1\\
\midrule
ANTERIOR INFARCT                     & 2         & 0.908     & 1.000             & 0.906             & 0.408    & 0.750                    & 0.999                    & 0.625           \\
ATRIAL FIBRILLATION                  & 59        & 0.996     & 1.000             & 0.972             & 0.866    & 0.798                    & 0.981                    & 0.796           \\
ATRIAL FLUTTER                       & 19        & 0.993     & 1.000             & 0.986             & 0.759    & 0.568                    & 0.990                    & 0.514           \\
ATRIAL-PACED RHYTHM                  & 3         & 0.991     & 1.000             & 0.988             & 0.500    & 0.833                    & 0.995                    & 0.657           \\
INCOMPLETE RIGHT BUNDLE BRANCH BLOCK & 4         & 0.893     & 1.000             & 0.698             & 0.506    & 0.563                    & 0.977                    & 0.299           \\
INFERIOR INFARCT                     & 1         & 0.998     & 1.000             & 0.998             & 0.667    & 1.000                    & 0.997                    & 0.725           \\
LATERAL INFARCT                      & 2         & 0.998     & 1.000             & 0.998             & 0.667    & 0.250                    & 0.999                    & 0.250           \\
LEFT BUNDLE BRANCH BLOCK             & 11        & 1.000     & 1.000             & 0.998             & 0.909    & 0.600                    & 0.932                    & 0.371           \\
NORMAL SINUS RHYTHM                  & 364       & 0.969     & 0.934             & 0.926             & 0.952    & 0.916                    & 0.862                    & 0.930           \\
PREMATURE ATRIAL COMPLEXES           & 19        & 0.997     & 1.000             & 0.963             & 0.679    & 0.579                    & 0.989                    & 0.613           \\
PREMATURE VENTRICULAR COMPLEXES      & 40        & 0.981     & 0.975             & 0.889             & 0.600    & 0.850                    & 0.996                    & 0.898           \\
RIGHT AXIS DEVIATION                 & 12        & 0.996     & 1.000             & 0.992             & 0.800    & 0.594                    & 0.979                    & 0.489           \\
RIGHT BUNDLE BRANCH BLOCK            & 50        & 0.984     & 0.940             & 0.969             & 0.847    & 0.775                    & 0.970                    & 0.762           \\
SINUS BRADYCARDIA                    & 41        & 0.995     & 1.000             & 0.989             & 0.938    & 0.717                    & 0.994                    & 0.791           \\
SINUS RHYTHM                         & 191       & 0.970     & 0.970             & 0.926             & 0.971    & 0.916                    & 0.862                    & 0.930           \\
SINUS TACHYCARDIA                    & 58        & 0.996     & 1.000             & 0.945             & 0.683    & 0.821                    & 0.990                    & 0.833           \\
VENTRICULAR TACHYCARDIA              & 9         & 0.903     & 1.000             & 0.882             & 0.635    & 0.875                    & 0.996                    & 0.678           \\
VENTRICULAR-PACED RHYTHM             & 20        & 0.988     & 0.950             & 0.940             & 0.559    & 0.725                    & 0.997                    & 0.756           \\
WITH 1ST DEGREE AV BLOCK             & 15        & 0.864     & 0.667             & 0.831             & 0.287    & 0.633                    & 0.962                    & 0.560           \\
WITH SINUS ARRHYTHMIA                & 6         & 0.976     & 1.000             & 0.945             & 0.308    & 0.542                    & 0.979                    & 0.319          \\
\bottomrule
\end{tabular}
\end{table*}


\begin{figure*}[h!]
    \centering
    \includegraphics[width=\textwidth]{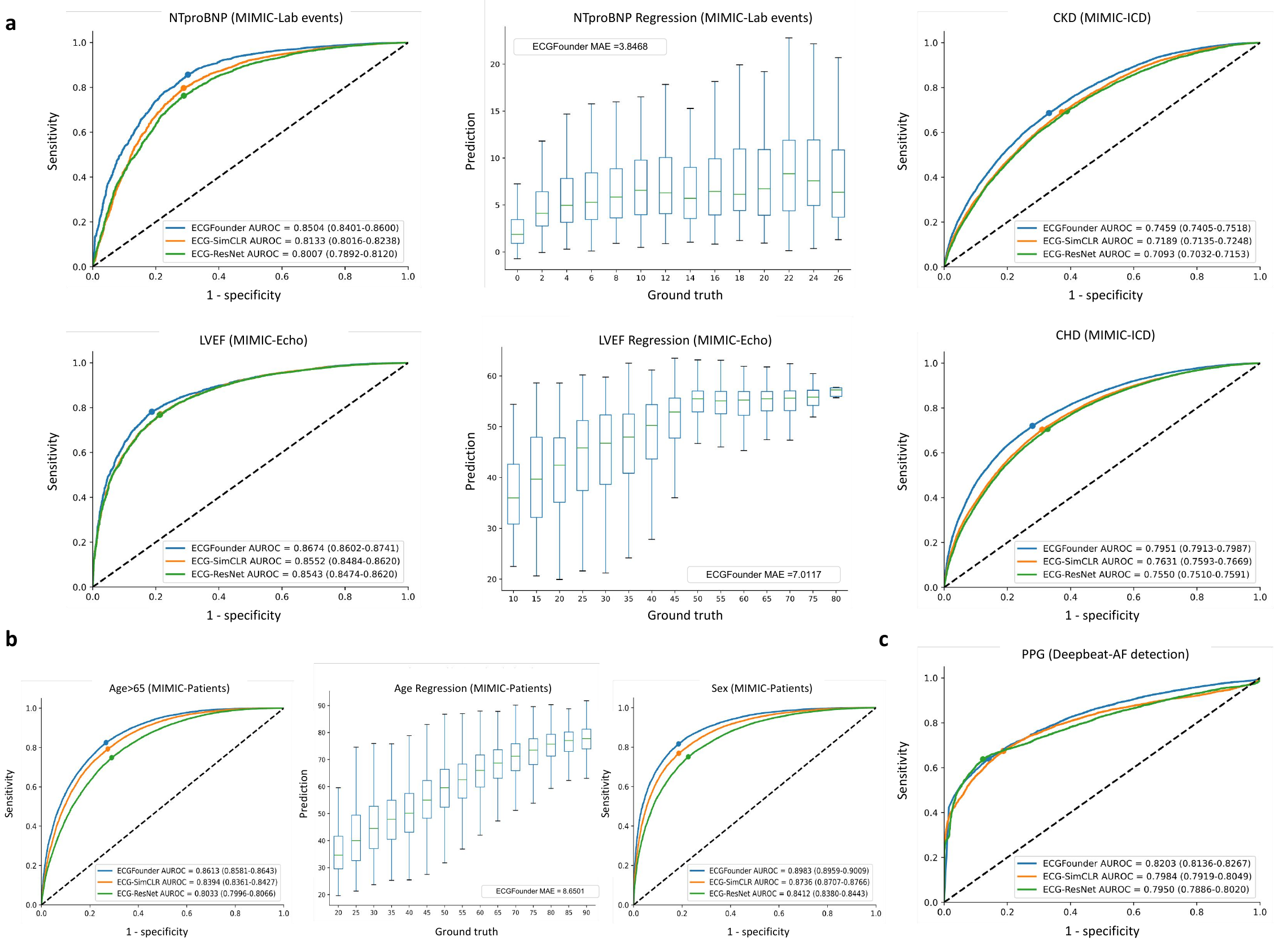} 
    \caption{(a) Results of NTproBNP classification, NTproBNP regression, CKD classification, LVEF classification, LVEF regression and CHD classification tasks of ECGFounder and other baseline models for clinical events detection tasks. (b) Results of age classification over 65 years old, age regression and sex classification tasks of ECGFounder and other baseline models for demographics detection tasks. (c) Results of PPG atrial fibrillation detection tasks.}
    \label{fig:ft_results}
\end{figure*}

We next validated the performance of ECGFounder in transfer learning. Our model was fine-tuned and evaluated on six downstream tasks using supervised learning on the MIMIC-IV-ECG dataset, resulting in both a 12-lead ECG model and a single-lead ECG model. The fine-tuning results are shown in Figures \ref{fig:ft_results}. As shown, ECGFounder outperforms the baseline methods in every downstream task. Specifically, it achieves 2 to 3 percentage points higher performance than ECG-SimCLR and 4 to 6 points higher than ECG-ResNet in age, sex, NTproBNP, CKD, and CHD detection. Additionally, to compare with the previously published CKD study \cite{holmstromDeepLearningbasedElectrocardiographic2023}, we created an independent test set consisting of individual patients from the MIMIC-IV-ECG dataset for validation. As shown in the results (Table \ref{tab:ckd}), our model surpasses the previously studied model in all six classifications for both 12-lead and single-lead ECGs.

\begin{table*}[]
\caption{Performance of other ECG deep learning model and ECGFounder on external test set CODE-test and PTB-XL.\cite{ribeiroAutomaticDiagnosis12lead2020,wagnerPTBXLLargePublicly2020} In particular, for the baseline method S12L-ECG, the CODE-test is the internal test set.}
\label{tab:PTBXL}
\begin{adjustbox}{width=\linewidth,center}
\begin{tabular}{ll|ccc|ccc|ccc|ccc}
\toprule
{Dataset} & {} & \multicolumn{12}{c}{CODE-test}
\\ \midrule
{Models} & {} & \multicolumn{3}{c}{\textbf{ECGFounder}} & \multicolumn{3}{c}{S12L-ECG(\textbf{Internal})\cite{ribeiroAutomaticDiagnosis12lead2020}} & \multicolumn{3}{c}{CTN\cite{natarajan2020wide}} & \multicolumn{3}{c}{ECG-SE-ResNet\cite{zhu2020classification}}
\\ \midrule
     & Count      & AUC   & Sens & Spec & AUC   & Sens & Spec & AUC   & Sens & Spec & AUC   & Sens & Spec \\
\midrule
SINUS BRADYCARDIA   & 16      & \textbf{0.967} & \textbf{1.000}       & 0.955       & 0.955 & 0.938       & \textbf{0.996 }      & 0.965 & 0.987       & 0.942       & 0.932 & 0.995       & 0.937       \\
ATRIAL FIBRILLATION  & 13     & \textbf{0.999} & \textbf{1.000}       & 0.996       & 0.963 & 0.769       & \textbf{1.000}       & 0.966 & 0.972       & 0.969       & 0.976 & 0.980       & 0.970       \\
SINUS TACHYCARDIA    & 37     & \textbf{0.989} & \textbf{0.974}       & 0.970       & 0.977 & 0.973       & \textbf{0.997  }     & 0.972 & 0.958       & 0.943       & 0.976 & 0.950       & 0.957       \\
RIGHT BUNDLE BRANCH BLOCK & 34 & 0.989 & 0.971       & 0.971       & \textbf{0.995} & \textbf{1.000 }     & \textbf{0.995}       & 0.989 & 0.946       & 0.986       & 0.965 & 0.949       & 0.947       \\
LEFT BUNDLE BRANCH BLOCK & 30 & 0.998 & \textbf{1.000}      & 0.996       & \textbf{1.000} & \textbf{1.000}       & \textbf{1.000}       & 0.961 & 0.988       & 0.942       & 0.971 & 0.949       & 0.976       \\
WITH 1ST DEGREE AV BLOCK & 28 & 0.949 & 0.864       & 0.957       & \textbf{0.989} & \textbf{0.929}       & \textbf{0.995}       & 0.930 & 0.862       & 0.925       & 0.958 & 0.856       & 0.922  
\\ \midrule
{Dataset} & {} & \multicolumn{12}{c}{PTB-XL}
\\ \midrule
{Models} & {} & \multicolumn{3}{c}{\textbf{ECGFounder}} & \multicolumn{3}{c}{S12L-ECG\cite{ribeiroAutomaticDiagnosis12lead2020}} & \multicolumn{3}{c}{CTN\cite{natarajan2020wide}} & \multicolumn{3}{c}{ECG-SE-ResNet\cite{zhu2020classification}}
\\ \midrule
ANTERIOR INFARCT                   & 360       & 0.635 & 0.816 & 0.447 & /     & /     & /     & /     & /     & /     & /     & /     & /     \\
ANTEROLATERAL INFARCT              & 297       & 0.945 & 0.837 & 0.918 & /     & /     & /     & /     & /     & /     & /     & /     & /     \\
ANTEROSEPTAL INFARCT               & 2415      & 0.945 & 0.891 & 0.882 & /     & /     & /     & /     & /     & /     & /     & /     & /     \\
ATRIAL FIBRILLATION                & 1514      & \textbf{0.993} & \textbf{0.969} & 0.982 & 0.925 & 0.776 & \textbf{0.989} & 0.953 & 0.935 & 0.915 & 0.950 & 0.946 & 0.907 \\
ATRIAL FLUTTER                     & 73        & 0.993 & 0.973 & 0.969 & /     & /     & /     & 0.948 & 0.962 & 0.937 & 0.941 & 0.966 & 0.956 \\
ELECTRONIC ATRIAL PACEMAKER        & 298       & 0.909 & 0.935 & 0.745 & /     & /     & /     & /     & /     & /     & /     & /     & /     \\
INFERIOR INFARCT                   & 2726      & 0.851 & 0.825 & 0.732 & /     & /     & /     & /     & /     & /     & /     & /     & /     \\
LATERAL INFARCT                    & 1066      & 0.916 & 0.819 & 0.859 & /     & /     & /     & /     & /     & /     & /     & /     & /     \\
LEFT ANTERIOR FASCICULAR BLOCK     & 1657      & \textbf{0.972} & \textbf{0.922} & \textbf{0.919} & /     & /     & /     & 0.965 & 0.882 & 0.914 & 0.968 & 0.819 & 0.901 \\
LEFT ATRIAL ENLARGEMENT            & 427       & 0.799 & 0.742 & 0.704 & /     & /     & /     & /     & /     & /     & /     & /     & /     \\
LEFT BUNDLE BRANCH BLOCK           & 539       & 0.989 & 0.969 & 0.965 & 0.983 & 0.963 & 0.991 & 0.912 & 0.932 & 0.933 & 0.931 & 0.917 & 0.930 \\
LEFT POSTERIOR FASCICULAR BLOCK    & 187       & 0.840 & 0.605 & 0.904 & /     & /     & /     & /     & /     & /     & /     & /     & /     \\
LEFT VENTRICULAR HYPERTROPHY       & 2419      & 0.900 & 0.860 & 0.788 & /     & /     & /     & /     & /     & /     & /     & /     & /     \\
LOW VOLTAGE QRS                    & 182       & 0.581 & \textbf{0.690} & 0.423 & /     & /     & /     & \textbf{0.605} & 0.605 & \textbf{0.814} & 0.593 & 0.655 & 0.741 \\
N-SPECIFIC INTRAVENTRICULAR BLOCK & 789       & 0.766 & 0.339 & \textbf{}0.932 & /     & /     & /     & \textbf{0.788} & 0.487 & \textbf{0.906} & 0.776 & \textbf{0.699} & 0.793 \\
NORMAL ECG                         & 9857      & 0.887 & 0.952 & 0.639 & /     & /     & /     & /     & /     & /     & /     & /     & /     \\
PREMATURE VENTRICULAR COMPLEXES    & 1146      & 0.987 & 0.965 & 0.959 & /     & /     & /     & 0.963 & 0.955 & 0.921 & 0.937 & 0.939 & 0.880 \\
QT HAS LENGTHENED                  & 119       & 0.931 & 0.897 & 0.827 & /     & /     & /     & 0.930 & 0.881 & 0.957 & 0.930 & 0.901 & 0.930 \\
RIGHT ATRIAL ENLARGEMENT           & 99        & 0.959 & 0.869 & 0.915 & /     & /     & /     & /     & /     & /     & /     & /     & /     \\
RIGHT BUNDLE BRANCH BLOCK          & 1155      & \textbf{0.976} & \textbf{0.925} & 0.925 & 0.967 & 0.911 & \textbf{0.976} & 0.932 & 0.922 & 0.931 & 0.946 & 0.842 & 0.912 \\
RIGHT VENTRICULAR HYPERTROPHY      & 136       & 0.913 & 0.833 & 0.859 & /     & /     & /     & /     & /     & /     & /     & /     & /     \\
SEPTAL INFARCT                     & 1423      & 0.947 & 0.894 & 0.880 & /     & /     & /     & /     & /     & /     & /     & /     & /     \\
SINUS BRADYCARDIA                  & 1159      & 0.950 & \textbf{0.972} & 0.955 & 0.967 & 0.938 & \textbf{0.989} & \textbf{0.996} & 0.966 & 0.946 & 0.983 & 0.935 & 0.963 \\
SINUS RHYTHM                       & 16785     & 0.923 & 0.936 & 0.751 & /     & /     & /     & 0.977 & 0.942 & 0.920 & 0.963 & 0.926 & 0.913 \\
SINUS TACHYCARDIA                  & 826       & \textbf{0.994} &\textbf{0.976}  & 0.987 & 0.988 & 0.946 & \textbf{0.996} & 0.986 & 0.957 & 0.968 & 0.979 & 0.932 & 0.944 \\
SUPRAVENTRICULAR TACHYCARDIA       & 27        & 0.995 & 0.976 & 0.959 & /     & /     & /     & /     & /     & /     & /     & /     & /     \\
VENTRICULAR TACHYCARDIA            & 24        & 0.987 & 0.905 & 0.993 & /     & /     & /     & /     & /     & /     & /     & /     & /     \\
WITH 1ST DEGREE AV BLOCK           & 802       & 0.911 & 0.661 & 0.921 & \textbf{0.962} & \textbf{0.925} & \textbf{0.992} & 0.943 & 0.783 & 0.911 & 0.930 & 0.735 & 0.893 \\
WITH QRS WIDENING                  & 45        & 0.618 & 0.414 & 0.742 & /     & /     & /     & /     & /     & /     & /     & /     & /     \\
WOLFF-PARKINSON-WHITE              & 18        & 0.919 & 0.747 & 0.982 & /     & /     & /     & /     & /     & /     & /     & /     & /     \\
\bottomrule
\end{tabular}
\end{adjustbox}
\end{table*}

\begin{table*}
\caption{Performance of other ECG deep learning model and ECGFounder on chronic kidney disease ECG test set. For ECGFounder, this is internal validation; for the Holmstrom's model, this is external validation.}
\label{tab:ckd}

\begin{tabular}{ll|cc|cc}

\toprule

{Models} & {} & \multicolumn{2}{c}{\textbf{ECGFounder}} & \multicolumn{2}{c}{Holmstrom et al.\cite{holmstromDeepLearningbasedElectrocardiographic2023}}
\\ \midrule
Class & Count & 12$-$lead AUC & 1$-$lead AUC & 12$-$lead AUC & 1$-$lead AUC \\ \midrule
Any-stage chronic kidney disease      &  13,990              & \textbf{0.746}       & \textbf{0.707}      & 0.639       & 0.622      \\
Mild chronic kidney disease &514    & \textbf{0.590}       & \textbf{0.590}      & 0.576       & 0.549      \\
Moderate to severe chronic kidney disease  &7075          & \textbf{0.713}     &\textbf{0.695}      & 0.661       & 0.632      \\
End stage renal disease        &6401            &\textbf{0.795}      &\textbf{0.725}      & 0.653       & 0.634   \\
\bottomrule
\end{tabular}
\end{table*}

\section{Discussion}
In this study, we have developed and demonstrated the generalizability and robust diagnostic capabilities of ECGFounder, a universal foundation model for electrocardiogram analysis. Trained on the largest ECG dataset to date, HEEDB, which encompasses over ten million ECGs from more than one million unique patients across 150 primary diagnostic categories including normal ECGs, arrhythmias, conduction blocks, myocardial infarctions, and cardiac hypertrophy. We developed and validated a deep learning model that consistently outperforms other ECG models. Our findings on both internal and external test sets highlight the substantial clinical diagnostic value and generalizability of our model. Furthermore, we enhanced the model's performance on single-lead ECGs through a novel data augmentation method based on the cardiac axis. The internal validation results for arrhythmia diagnosis using single-lead ECGs demonstrated exceptional performance, broadening the prospects for the model’s application in mobile health. Moreover, by leveraging a fine-tuned pre-trained model, ECGFounder effectively adapts to a wide range of downstream tasks, significantly enhancing the diagnosis of other diseases such as chronic kidney disease and coronary heart disease.

ECGFounder enhances diagnostic performance by learning to identify ECG features associated with cardiovascular diseases, which are typically diagnosed based on specific waveform patterns and rhythm characteristics like the elevated ST segments of myocardial infarction and the irregular fluctuations of atrial fibrillation. These features involve anomalies in cardiac electrical activity, appearing significantly different from normal ECG waveforms. Upon training, ECGFounder can recognize these disease-related waveform patterns and rhythms, accurately diagnosing corresponding cardiovascular conditions. As observed in Table \ref{tab:internal} and Figure \ref{fig:eval} in Supplement , ECGFounder matches or even exceeds the performance of cardiology experts on the internal review test sets. Furthermore, ECGFounder’s sensitivity exceeds that of the average cardiology expert, indicating its ability to more accurately capture subtle signs of cardiovascular diseases that may be overlooked by human experts.

Previous research has demonstrated that deep learning models can support clinical ECG analysis and achieve good performance.\cite{hannunCardiologistlevelArrhythmiaDetection2019,hughesPerformanceConvolutionalNeural2021,zhuAutomaticMultilabelElectrocardiogram2020} However, existing models lack a universal clinical diagnostic capability for ECGs. Firstly, the training datasets used by existing models are not large or diverse enough, which can lead to overfitting and poor performance on new data, thus limiting their generalizability.\cite{schlapfer2017computer} Additionally, the lack of demographic diversity in the training datasets means these models may perform poorly for certain demographic groups. This could lead to biases when the models are applied to data from different demographic backgrounds, affecting the fairness and accuracy of the models. Secondly, the labels for most model training datasets are derived from manual annotations by cardiology experts, which is time-consuming and labor-intensive. This limits the number of ECGs available for training. Also, because cardiology experts typically use a unified annotation system, the richness of the dataset labels is not very high, often only including common ECG abnormalities and omitting many important but rare diagnostic labels.\cite{strodthoffDeepLearningECG2021} Thirdly, due to the training methods used, these models do not support both 12-lead and single-lead ECGs. To address these issues, we propose a foundation model for ECGs that supports both 12-lead and single-lead ECGs. In Tables \ref{tab:PTBXL}, we observed that ECGFounder ranks first in average performance across various external tests. The other baseline models, including S12L-ECG, CTN, and ECG-SE-ResNet, have previously achieved the best performance in AI-ECG models.\cite{natarajan2020wide,zhu2020classification,ribeiroAutomaticDiagnosis12lead2020} S12L-ECG was trained using supervised learning on the CODE-ECG dataset, which includes 2 million ECGs with 6 common types of arrhythmia diagnoses.\cite{ribeiroAutomaticDiagnosis12lead2020} CTN and ECG-SE-ResNet were trained on the PhysioNet Challenge-2020 dataset, which includes 60,000 ECGs covering 27 common ECG diagnoses.\cite{zhu2020classification,natarajan2020wide} In this paper, we demonstrate that by training on a larger and more diverse ECG dataset, a scalable foundation model can be developed to further improve the diagnosis of cardiovascular diseases and surpass previous baseline models.

Despite the effectiveness of ECGFounder in detecting various cardiovascular diseases, there are still challenges to be addressed. Firstly, as most data used to develop ECGFounder come from U.S. cohorts, this limits the diversity and representativeness of the data. Different regions of the world may have unique ECG characteristics, such as race and regional-specific rhythm variations, requiring the model to handle data from diverse backgrounds and populations. Secondly, although ECG foundation models can provide high-accuracy diagnostic results, their decision-making process is often a 'black box,' which might encounter trust and acceptance issues in clinical applications. Therefore, developing explainable AI models, allowing doctors to understand the model's decision-making process, is key to advancing the use of ECG AI models. Lastly, some clinically relevant information, such as patients' lifestyles and medical histories, which could serve as effective covariates in cardiovascular disease research, have not yet been included in the model. We suggest that future work should involve incorporating a larger volume of ECG data from different regions and ethnicities, adding demographic information of patients as model inputs, and developing more explainable AI models, enabling doctors to better understand the decision processes and outcomes of the models. At the same time, some of the latest natural language processing methods have shown great potential in handling ECG annotations from experts, allowing for a more effective use of the clinical knowledge embedded in these annotations for learning.\cite{li2024frozen}

\section{Conclusion}
In this work, we proposed and validated the efficacy of ECGFounder in adapting to a wide range of cardiovascular diagnostic applications, demonstrating its high performance and versatility across various downstream tasks as a foundational ECG model. By overcoming the limitations related to ECG data and labeling quality and diversity, as well as training methods, our ECG foundation model confirms its potential to transform the standard of care in cardiology and to provide real-time, accurate cardiac assessments in diverse clinical settings. We have provided open access to the model, as well as the code and data used to train it under an open access license. In this way, we invite the research community to continue build upon our model, and help advance the state-of-the-art.

\section*{Acknowledgements}

This document is the result of the research project funded by the National Science Foundation and Emory University via an unrestricted gift. 

Retrospective analysis of data for this project was conducted with waiver of informed consent under approved IRB protocols (BIDMC: 2022P000417; MGH: 2013P001024).

Funding: Dr. Westover was supported by grants from the NIH (RF1AG064312, RF1NS120947, R01AG073410, R01HL161253, R01NS126282, R01AG073598, R01NS131347, R01NS130119), and NSF (2014431). Dr. Shenda Hong was supported by the National Natural Science Foundation of China (62102008), Clinical Medicine Plus X - Young Scholars Project of Peking University, the Fundamental Research Funds for the Central Universities (PKU2024LCXQ030). For this work, Dr. Clifford was partially supported by the National Institute of Biomedical Imaging and Bioengineering (NIBIB) under NIH award number R01EB030362. 


Disclosures: Dr. Westover is a co-founder, scientific advisor, and consultant to Beacon Biosignals and has a personal equity interest in the company. Dr. Clifford holds significant stock in AliveCor.

The content is solely the responsibility of the authors and does not necessarily represent the official views of the National Institutes of Health, or the authors' current and past employers and funding bodies. 

We thank Dr. Xinxin Di, Dr. Kun Lu, Dr. Zhengkai Xue, Dr. Wenbo Dai, Dr. Jing Zhao, Dr. Hongqian Zhou for annotating the internal test set.

\bibliographystyle{unsrtnat}
\clearpage
\bibliography{ECGFounder}

\newpage

\section*{Supplement}
\setcounter{page}{0}
\pagenumbering{arabic}
\renewcommand*{\thepage}{P\arabic{page}}

\setcounter{figure}{0}
\renewcommand{\thefigure}{S\arabic{figure}}
\setcounter{table}{0}
\renewcommand{\thetable}{S\arabic{table}}

\subsection*{S1. Q\&A of the ECGFounder}
\paragraph{What is the ECG foundation model?}
The ECG foundation model is a large-scale, pre-trained neural network designed for tasks involving ECG (electrocardiogram) data. This model aims to capture and learn generalizable features from ECG signals, which can be fine-tuned or adapted for a wide range of downstream tasks like disease classification, arrhythmia detection,  or patient-specific anomaly identification.

In our study, we propose the ECGFounder, which is being developed using the Harvard-Emory ECG Dataset. The model leverages this diverse and comprehensive dataset to create a robust feature extractor that can later be fine-tuned or used in transfer learning scenarios for specific diagnostic or research objectives. The ultimate goal is to have a powerful, generalized model that can perform various ECG-related tasks with minimal additional training on specific datasets.

\paragraph{What is the use of ECG foundation model?}
We provide the medical AI community with a versatile ECG foundation model that serves as a comprehensive feature extractor and a highly adaptable base for transfer learning tasks. Compared to traditional models trained from scratch, this foundation model offers a generalized representation of ECG signals, enabling enhanced performance across diverse downstream tasks through fine-tuning. By leveraging large-scale pre-training on the Harvard-Emory ECG Dataset, the model captures robust and clinically relevant features, thus reducing the need for extensive labeled datasets for specific applications. This advancement holds significant potential to drive future innovations in AI-based ECG analysis, disease detection, and personalized healthcare solutions. And we offer an effective out-of-the-box solution and fine-tuning capabilities for the community to use it more easily.

\paragraph{How is the performance of the ECG foundation model?}
The ECG foundation model demonstrates strong performance through several tasks, showcasing its high accuracy and generalization capabilities due to pre-training on large dataset like the Harvard-Emory ECG Dataset. It achieves superior results across various downstream tasks, including arrhythmia classification, disease diagnosis, and demographics detection, with fine-tuning further enhancing performance compared to models trained from scratch. The model effectively captures important ECG features, enabling precise differentiation of cardiac conditions with minimal domain-specific training. Additionally, its pre-trained architecture allows for efficient transfer learning, requiring fewer labeled samples and training epochs, thereby reducing time and computational costs. Its adaptability extends beyond disease classification to tasks like demographics and laboratory  measurement detection, highlighting its versatility. Validation results consistently demonstrate that the foundation model outperforms baseline ECG models, achieving higher AUC, making it a robust and efficient tool for advancing AI applications in cardiology and beyond. The table \ref{tab:class_performance} and \ref{tab:reg_performance} show the performance of ECG foundation model.

\begin{table}[h]
\centering
\caption{Performance of classification tasks of the ECG foundation model.}
\label{tab:class_performance}
\begin{tabular}{lp{4cm}ccc}
\toprule
\textbf{Dataset} & \textbf{Task type} & \textbf{ECGFounder} & \textbf{ECG-SimCLR} & \textbf{ECG-ResNet} \\ \midrule
MIMIC-ECG-NTproBNP       & Laboratory  measurement                     & \textbf{0.8504}          & 0.8133           & 0.8007         \\
MIMIC-ECG-CKD         & Disease diagnosis                         & \textbf{0.7459}         & 0.7189          & 0.7093         \\
MIMIC-ECG-LVEF        & Medical image metric                         & \textbf{0.8674}          & 0.8552           & 0.8543           \\
MIMIC-ECG-CHD      & Disease diagnosis                            & \textbf{0.7951}         & 0.7631          & 0.7550         \\
MIMIC-ECG-Age      & Demographics                            & \textbf{0.8613}         & 0.8394          & 0.8033         \\
MIMIC-ECG-Sex      & Demographics                            & \textbf{0.8983}         & 0.8736          & 0.8412         \\
\midrule
Deepbeat(PPG)        & PPG AF detection                         & \textbf{0.8203}          & 0.7984            & 0.7950         \\
\bottomrule
\end{tabular}
\end{table}

\begin{table}[h]
\centering
\caption{Performance of regression tasks of the ECG foundation model.}
\label{tab:reg_performance}
\begin{tabular}{lp{4cm}ccc}
\toprule
\textbf{Dataset} & \textbf{Task type} & \textbf{ECGFounder} & \textbf{ECG-SimCLR} & \textbf{ECG-ResNet} \\ \midrule
MIMIC-ECG-NTproBNP   & Laboratory  measurement                             & \textbf{3.8468}          & 4.8873           & 4.9893         \\
MIMIC-ECG-LVEF       & Medical image metric                             & \textbf{7.0117}          & 7.0579           & 7.9514         \\
MIMIC-ECG-Age        & Demographics                             & \textbf{8.6501}         & 9.0153          & 9.4004         \\
\bottomrule
\end{tabular}
\end{table}

\paragraph{How can researchers use the ECG Foundation Model?}
The trained model and labeled data will be publicly released upon publication through the Brain Data Science Platform (bdsp.io), huggingface and PhysioNet.

\newpage

\subsection*{S2. Details of Datasets}

\subsubsection*{S2.1. Details of pre-training dataset}
Table \ref{tb:data} is a supplement to Figure 1.c in the main text. The table compares the size and label types of ECG datasets used in multiple existing works. It can be observed that Harvard-Emory ECG Database(HEEDB) has far more data sizes and label types than other datasets. And table 2 describes the label types and corresponding quantities of all HEEDB datasets.

\begin{table}[H]
\centering
\caption{Comparison with ECG dataset of HEEDB and existing works.}
\begin{tabular}{lrrrr} 
\toprule
     & ECGs & Patients & Classes & Lead \\ \midrule
Stanford-ECG \cite{hannunCardiologistlevelArrhythmiaDetection2019}   & 91,232  & 53,549 & 12  & 1  \\
PTB-XL \cite{wagnerPTBXLLargePublicly2020}   &  21,837   & 18,885  &  71  & 12        \\
PhysioNet Challenge 2020 \cite{ErickAPerezAlday_2020_12_29_Classificationof} & 66,361   & Unknown & 27   & 12       \\
CODE \cite{ribeiroAutomaticDiagnosis12lead2020}  & 2,322,513  & 1,676,384 & 6  & 12   \\ 
Tongji-ECG \cite{zhuAutomaticMultilabelElectrocardiogram2020}  & 180,112  & 70,692 & 21  & 12   \\
UCSF-ECG \cite{hughesPerformanceConvolutionalNeural2021}  & 992,748  & 365,009 & 38  & 12   \\
\hline
HEEDB (Ours) \cite{koscovaHarvardEmoryECGDatabase2024}    & 10,771,552 & 1,818,247 & 150    & 12       \\ \bottomrule
\label{tb:data}
\end{tabular}
\end{table}

\begin{longtable}{|l|l|}
\caption{Filtered 150 meaningful label list of HEEDB development set.} \\
\hline
\textbf{Label} & \textbf{Count} \\ \hline
\endfirsthead

\multicolumn{2}{c}{{\bfseries \tablename\ \thetable{} -- continued from previous page}} \\
\hline
\textbf{Label} & \textbf{Count} \\ \hline
\endhead

\hline \multicolumn{2}{r}{{Continued on next page}} \\
\endfoot

\hline
\endlastfoot
\hline
ABNORMAL ECG                                                         & 5916870 \\ \hline
NORMAL SINUS RHYTHM                                                  & 5440634 \\ \hline
NORMAL ECG                                                           & 1912774 \\ \hline
SINUS RHYTHM                                                         & 1583298 \\ \hline
SINUS BRADYCARDIA                                                    & 1188526 \\ \hline
ATRIAL FIBRILLATION                                                  & 1108003 \\ \hline
SINUS TACHYCARDIA                                                    & 964612  \\ \hline
OTHERWISE NORMAL ECG                                                 & 927256  \\ \hline
LEFT AXIS DEVIATION                                                  & 880984  \\ \hline
PREMATURE VENTRICULAR COMPLEXES                                      & 834752  \\ \hline
BORDERLINE ECG                                                       & 697210  \\ \hline
RIGHT BUNDLE BRANCH BLOCK                                            & 684092  \\ \hline
SEPTAL INFARCT                                                       & 662652  \\ \hline
LEFT ATRIAL ENLARGEMENT                                              & 637167  \\ \hline
NONSPECIFIC T WAVE ABNORMALITY                                       & 578612  \\ \hline
LOW VOLTAGE QRS                                                      & 528770  \\ \hline
PREMATURE ATRIAL COMPLEXES                                           & 498470  \\ \hline
ANTERIOR INFARCT                                                     & 380691  \\ \hline
INCOMPLETE RIGHT BUNDLE BRANCH BLOCK                                 & 377899  \\ \hline
PREMATURE SUPRAVENTRICULAR COMPLEXES                                 & 333798  \\ \hline
LEFT BUNDLE BRANCH BLOCK                                             & 318791  \\ \hline
NONSPECIFIC T WAVE ABNORMALITY NOW EVIDENT IN                        & 299977  \\ \hline
NONSPECIFIC T WAVE ABNORMALITY NO LONGER EVIDENT IN                  & 275433  \\ \hline
T WAVE INVERSION NOW EVIDENT IN                                      & 264182  \\ \hline
LATERAL INFARCT                                                      & 263295  \\ \hline
NONSPECIFIC ST ABNORMALITY                                           & 260714  \\ \hline
LEFT VENTRICULAR HYPERTROPHY                                         & 257036  \\ \hline
T WAVE INVERSION NO LONGER EVIDENT IN                                & 251761  \\ \hline
WITH RAPID VENTRICULAR RESPONSE                                      & 243049  \\ \hline
QT HAS SHORTENED                                                     & 221554  \\ \hline
QT HAS LENGTHENED                                                    & 216377  \\ \hline
FUSION COMPLEXES                                                     & 198828  \\ \hline
ATRIAL FLUTTER                                                       & 198007  \\ \hline
MARKED SINUS BRADYCARDIA                                             & 183847  \\ \hline
WITH SINUS ARRHYTHMIA                                                & 182984  \\ \hline
NONSPECIFIC ST AND T WAVE ABNORMALITY                                & 177718  \\ \hline
LEFT ANTERIOR FASCICULAR BLOCK                                       & 154028  \\ \hline
RIGHT AXIS DEVIATION                                                 & 153561  \\ \hline
ECTOPIC ATRIAL RHYTHM                                                & 151104  \\ \hline
UNDETERMINED RHYTHM                                                  & 148020  \\ \hline
ANTEROSEPTAL INFARCT                                                 & 136656  \\ \hline
RIGHTWARD AXIS                                                       & 130331  \\ \hline
ST NOW DEPRESSED IN                                                  & 128118  \\ \hline
WITH SHORT PR                                                        & 126460  \\ \hline
WITH MARKED SINUS ARRHYTHMIA                                         & 124492  \\ \hline
ST NO LONGER DEPRESSED IN                                            & 113896  \\ \hline
INVERTED T WAVES HAVE REPLACED NONSPECIFIC T WAVE ABNORMALITY IN     & 109365  \\ \hline
NON-SPECIFIC CHANGE IN ST SEGMENT IN                                 & 109261  \\ \hline
NONSPECIFIC T WAVE ABNORMALITY HAS REPLACED INVERTED T WAVES IN      & 109217  \\ \hline
JUNCTIONAL RHYTHM                                                    & 108731  \\ \hline
ELECTRONIC ATRIAL PACEMAKER                                          & 107890  \\ \hline
ABERRANT CONDUCTION                                                  & 103597  \\ \hline
ELECTRONIC VENTRICULAR PACEMAKER                                     & 96125   \\ \hline
T WAVE INVERSION LESS EVIDENT IN                                     & 94837   \\ \hline
ANTEROLATERAL INFARCT                                                & 92527   \\ \hline
WITH REPOLARIZATION ABNORMALITY                                      & 91043   \\ \hline
RSR' OR QR PATTERN IN V1 SUGGESTS RIGHT VENTRICULAR CONDUCTION DELAY & 90752   \\ \hline
T WAVE INVERSION MORE EVIDENT IN                                     & 90476   \\ \hline
WIDE QRS RHYTHM                                                      & 89366   \\ \hline
WITH PREMATURE VENTRICULAR OR ABERRANTLY CONDUCTED COMPLEXES         & 88985   \\ \hline
RIGHT ATRIAL ENLARGEMENT                                             & 64829   \\ \hline
INFERIOR INFARCT                                                     & 64044   \\ \hline
INCOMPLETE LEFT BUNDLE BRANCH BLOCK                                  & 63540   \\ \hline
VOLTAGE CRITERIA FOR LEFT VENTRICULAR HYPERTROPHY                    & 62822   \\ \hline
OR DIGITALIS EFFECT                                                  & 62191   \\ \hline
BIFASCICULAR BLOCK                                                   & 59790   \\ \hline
ST NO LONGER ELEVATED IN                                             & 58002   \\ \hline
WITH SLOW VENTRICULAR RESPONSE                                       & 56777   \\ \hline
ST ELEVATION NOW PRESENT IN                                          & 56383   \\ \hline
PREMATURE ECTOPIC COMPLEXES                                          & 55818   \\ \hline
LEFT POSTERIOR FASCICULAR BLOCK                                      & 55139   \\ \hline
T WAVE AMPLITUDE HAS DECREASED IN                                    & 44131   \\ \hline
WITH A COMPETING JUNCTIONAL PACEMAKER                                & 42770   \\ \hline
RIGHT SUPERIOR AXIS DEVIATION                                        & 40107   \\ \hline
BIATRIAL ENLARGEMENT                                                 & 40051   \\ \hline
VENTRICULAR-PACED RHYTHM                                             & 39485   \\ \hline
ATRIAL-PACED RHYTHM                                                  & 39290   \\ \hline
T WAVE AMPLITUDE HAS INCREASED IN                                    & 38556   \\ \hline
WITH QRS WIDENING                                                    & 37944   \\ \hline
WITH 1ST DEGREE AV BLOCK                                             & 36157   \\ \hline
PROLONGED QT                                                         & 34564   \\ \hline
WITH PROLONGED AV CONDUCTION                                         & 34345   \\ \hline
RIGHT VENTRICULAR HYPERTROPHY                                        & 34095   \\ \hline
WITH QRS WIDENING AND REPOLARIZATION ABNORMALITY                     & 33211   \\ \hline
ATRIAL-SENSED VENTRICULAR-PACED RHYTHM                               & 32599   \\ \hline
AV SEQUENTIAL OR DUAL CHAMBER ELECTRONIC PACEMAKER                   & 31180   \\ \hline
PULMONARY DISEASE PATTERN                                            & 31111   \\ \hline
ACUTE MI / STEMI                                                     & 30735   \\ \hline
INFERIOR-POSTERIOR INFARCT                                           & 29775   \\ \hline
NONSPECIFIC INTRAVENTRICULAR CONDUCTION DELAY                        & 28423   \\ \hline
PREMATURE VENTRICULAR AND FUSION COMPLEXES                           & 26524   \\ \hline
IN A PATTERN OF BIGEMINY                                             & 25765   \\ \hline
AV DUAL-PACED RHYTHM                                                 & 25056   \\ \hline
SUPRAVENTRICULAR TACHYCARDIA                                         & 18945   \\ \hline
VENTRICULAR-PACED COMPLEXES                                          & 17459   \\ \hline
WIDE QRS TACHYCARDIA                                                 & 17029   \\ \hline
RSR' PATTERN IN V1                                                   & 16953   \\ \hline
ST LESS DEPRESSED IN                                                 & 15676   \\ \hline
VENTRICULAR TACHYCARDIA                                              & 15603   \\ \hline
EARLY REPOLARIZATION                                                 & 15177   \\ \hline
ST MORE DEPRESSED IN                                                 & 14426   \\ \hline
ANTEROLATERAL LEADS                                                  & 14209   \\ \hline
ELECTRONIC DEMAND PACING                                             & 14089   \\ \hline
RBBB AND LEFT ANTERIOR FASCICULAR BLOCK                              & 11593   \\ \hline
LATERAL INJURY PATTERN                                               & 10815   \\ \hline
BIVENTRICULAR PACEMAKER DETECTED                                     & 10686   \\ \hline
SUSPECT UNSPECIFIED PACEMAKER FAILURE                                & 10498   \\ \hline
WOLFF-PARKINSON-WHITE                                                & 10264   \\ \hline
WITH VENTRICULAR ESCAPE COMPLEXES                                    & 10240   \\ \hline
INFERIOR INJURY PATTERN                                              & 10033   \\ \hline
CONSIDER RIGHT VENTRICULAR INVOLVEMENT IN ACUTE INFERIOR INFARCT     & 9978    \\ \hline
ST ELEVATION HAS REPLACED ST DEPRESSION IN                           & 9585    \\ \hline
NONSPECIFIC INTRAVENTRICULAR BLOCK                                   & 9258    \\ \hline
MASKED BY FASCICULAR BLOCK                                           & 9074    \\ \hline
PEDIATRIC ECG ANALYSIS                                               & 8707    \\ \hline
BLOCKED                                                              & 8529    \\ \hline
WITH UNDETERMINED RHYTHM IRREGULARITY                                & 8365    \\ \hline
LEFTWARD AXIS                                                        & 7690    \\ \hline
WITH 2ND DEGREE SA BLOCK MOBITZ I                                    & 7380    \\ \hline
ACUTE                                                                & 6857    \\ \hline
ABNORMAL LEFT AXIS DEVIATION                                         & 6708    \\ \hline
WITH COMPLETE HEART BLOCK                                            & 6505    \\ \hline
NO P-WAVES FOUND                                                     & 6384    \\ \hline
ST LESS ELEVATED IN                                                  & 5406    \\ \hline
WITH RETROGRADE CONDUCTION                                           & 5186    \\ \hline
ST MORE ELEVATED IN                                                  & 4982    \\ \hline
JUNCTIONAL BRADYCARDIA                                               & 4651    \\ \hline
WITH VARIABLE AV BLOCK                                               & 4556    \\ \hline
ANTERIOR INJURY PATTERN                                              & 4412    \\ \hline
WITH JUNCTIONAL ESCAPE COMPLEXES                                     & 4077    \\ \hline
ACUTE MI                                                             & 3867    \\ \hline
ACUTE PERICARDITIS                                                   & 3745    \\ \hline
POSTERIOR INFARCT                                                    & 3657    \\ \hline
IDIOVENTRICULAR RHYTHM                                               & 3609    \\ \hline
WITH 2ND DEGREE SA BLOCK MOBITZ II                                   & 2912    \\ \hline
R IN AVL                                                             & 2723    \\ \hline
SINUS/ATRIAL CAPTURE                                                 & 2603    \\ \hline
AV DUAL-PACED COMPLEXES                                              & 2516    \\ \hline
INFEROLATERAL INJURY PATTERN                                         & 2204    \\ \hline
RBBB AND LEFT POSTERIOR FASCICULAR BLOCK                             & 1922    \\ \hline
ANTEROLATERAL INJURY PATTERN                                         & 1804    \\ \hline
ATRIAL-PACED COMPLEXES                                               & 1655    \\ \hline
WITH SINUS PAUSE                                                     & 1518    \\ \hline
BIVENTRICULAR HYPERTROPHY                                            & 1453    \\ \hline
ABNORMAL RIGHT AXIS DEVIATION                                        & 1314    \\ \hline
SUPRAVENTRICULAR COMPLEXES                                           & 1291    \\ \hline
WITH 2ND DEGREE AV BLOCK MOBITZ I                                    & 1152    \\ \hline
WITH 2:1 AV CONDUCTION                                               & 1080    \\ \hline
WITH AV DISSOCIATION                                                 & 1030    \\ \hline
MULTIFOCAL ATRIAL TACHYCARDIA                                        & 1016    \\ \hline
\end{longtable}

\newpage

\subsubsection*{S2.2. Details of fine-tuning dataset}

To evaluate the capability of transfer learning of ECGFounder, we developed 10 kinds of downstream datasets for fine-tuning.

For the MIMIC-ECG-NTproBNP dataset, we extracted data from the lab events section of MIMIC-IV-Hosp, obtaining NT-proBNP values matched with corresponding patients and admission IDs. According to clinically recognized laboratory standards, NT-proBNP values were categorized into two groups: normal and abnormal. This classification, along with the raw NT-proBNP values, was then matched to the corresponding ECGs of the patients for use in classification and regression tasks, respectively. Additionally, considering that abnormal NT-proBNP values can sometimes be excessively high, leading to non-convergence in the regression loss, we applied z-score normalization to these values. All normalized values were scaled to a range of 0 to 50.

For the MIMIC-ECG-CKD dataset, we extracted data from the diagnoses-ICD section of MIMIC-IV-Hosp, obtaining ICD codes matched with corresponding patients and admission IDs. Considering that the MIMIC database contains both ICD-9 and ICD-10 formats, we selected ICD-10 as the reference standard and converted all ICD codes to ICD-10. The ICD-10 codes for chronic kidney disease (CKD) include N182 (Chronic kidney disease, stage 2 [mild]), N183 (Chronic kidney disease, stage 3 [moderate]), N184 (Chronic kidney disease, stage 4 [severe]), N185 (Chronic kidney disease, stage 5), and N186 (End-stage renal disease). These codes were consolidated and treated collectively as indicators of CKD. If a patient has a positive diagnosis under any of these codes, they are considered to have CKD. This CKD classification information was then matched to the corresponding ECGs of the patients for use in the CKD classification task.

For the MIMIC-ECG-CHD dataset, we extracted ICD codes following the same procedure as described above. The ICD-10 codes for chronic heart disease (CHD) include all codes from I20 to I25. These codes were consolidated and treated collectively as indicators of CHD. If a patient has a positive diagnosis under any of these codes, they are considered to have CHD. This CHD classification information was then matched to the corresponding ECGs of the patients for use in the CHD classification task.

For the MIMIC-ECG-LVEF dataset, we extracted data from the discharge section of MIMIC-IV-Notes, obtaining echocardiogram notes and corresponding LVEF values matched with patients and admission IDs. According to clinically recognized echocardiogram diagnostic standards, an LVEF of 50\% or higher is considered normal, while an LVEF below 50\% is considered abnormal. This LVEF classification information, along with the raw LVEF values, was matched to the corresponding ECGs of the patients for use in both classification and regression tasks.

For the MIMIC-ECG-Age dataset, we extracted data from the patients section of MIMIC-IV-Hosp, obtaining age information matched with the corresponding patients. Ages were classified using a threshold of 65 years: individuals aged 65 or older were considered elderly, while those under 65 were not. This age classification, along with the raw age values, was matched to the corresponding ECGs of the patients for use in both classification and regression tasks.

For the MIMIC-ECG-Sex dataset, we extracted data from the patients section of MIMIC-IV-Hosp, obtaining sex information matched with the corresponding patients. This sex classification information was then matched to the corresponding ECGs of the patients for use in the sex classification task.

For the PPG Dataset, we utilized the DeepBeat-PPG dataset, which includes PPG data from 132 patients with atrial fibrillation (AF). This dataset provides publicly available validation and test sets. Here, we used the validation set as both the training and validation set, and the test set as a holdout test set. This approach ensures that patients do not overlap between the training and test sets, allowing for a fair performance comparison.

\begin{table}[H]
\centering
\caption{Details on pre-training and fine-tuning dataset split.}
\label{tab:data_split}
\begin{tabular}{lcccc}
\toprule
\textbf{Dataset} & \textbf{Task type} & \textbf{Train} & \textbf{Valid} & \textbf{Test} \\ \midrule
HEEDB(Pre-training)                 & classification                         & 6,683,587       & 835,448         & 834,926 \\ 
\midrule
MIMIC-ECG-NTproBNP       & classification                      & 47,084          & 5,884           & 5,893         \\
MIMIC-ECG-CKD         & classification                         & 213,926         & 26,749          & 26,741         \\
MIMIC-ECG-LVEF        & classification                         & 60,333          & 7,539           & 7,541           \\
MIMIC-ECG-CHD      & classification                            & 213,926         & 26,749          & 26,741         \\
MIMIC-ECG-Age      & classification                            & 640,027         & 80,015          & 80,004         \\
MIMIC-ECG-Sex      & classification                            & 640,027         & 80,015          & 80,004         \\
\midrule
MIMIC-ECG-NTproBNP   & regression                              & 47,084          & 5,884           & 5,893         \\
MIMIC-ECG-LVEF       & regression                              & 60,333          & 7,539           & 7,541         \\
MIMIC-ECG-Age        & regression                              & 640,027         & 80,015          & 80,004         \\
\midrule
Deepbeat(PPG)        & classification                         & 466,904          & 51,878            & 17,617         \\
\bottomrule
\end{tabular}
\end{table}

\subsubsection*{S2.3. Details of cardiologists}
Table \ref{tab:cardiologists} describes the personal information of 4 cardiologists participating in Cardiologist versus ECGFounder, including working year, gender, age and education level.

\begin{table}[h]
\centering
\caption{Working year, gender, age and education level of the cardiologists participating in the comparison with ECGFounder.}
\label{tab:cardiologists}
\begin{tabular}{lcccc}
\hline
Member                 & Working year & Gender & Age & Education level    \\ \hline
Student Cardiologist 1 & 2            &  Male       & 25  & M.D. Candidate \\ 
Student Cardiologist 2 & 3            &  Female     & 27  & M.D. Candidate \\ 
Expert Cardiologist 1  & 27           &  Female     & 50  & Master       \\
Expert Cardiologist 2  & 25           &  Male       & 48  & Master       \\ \hline
\end{tabular}
\end{table}

\newpage


\subsection*{S3. Details of Method}
\subsubsection*{S3.1. Details of model architecture}
\begin{figure}[hbp]
    \centering
    \includegraphics[width=\textwidth]{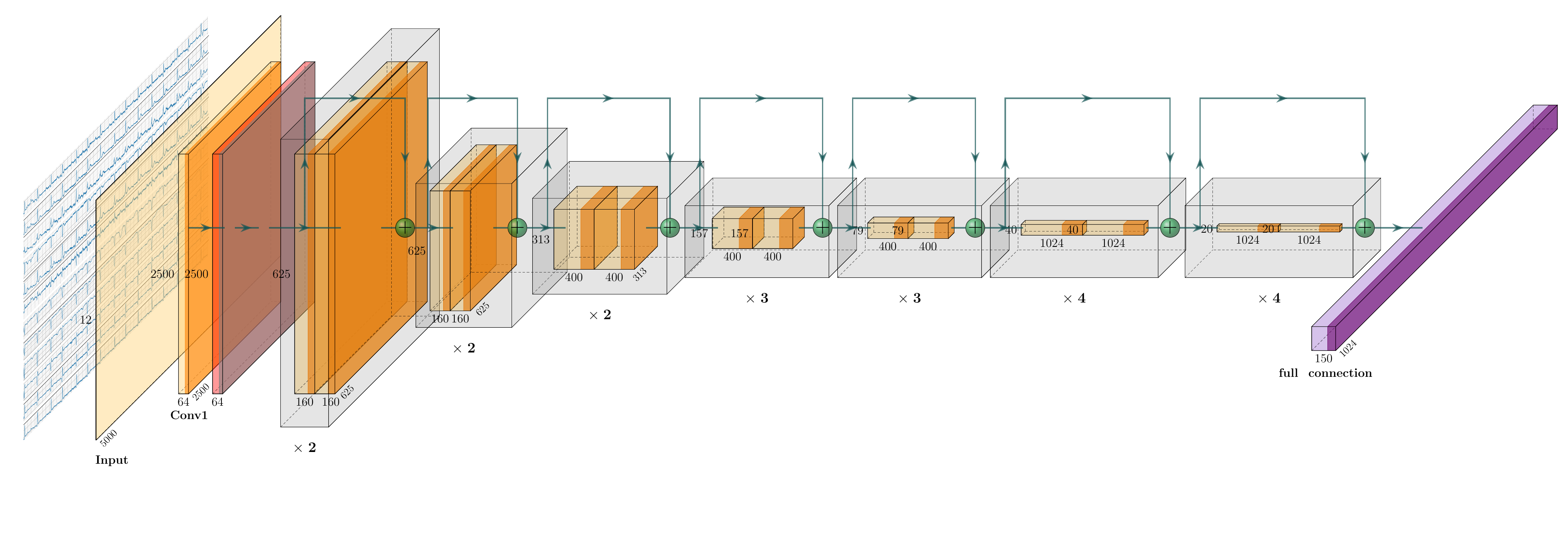} 
    \caption{The detailed architecture plot of the ECGFounder, indicating the dimensions of key components of the model.}
    \label{fig:ECG_model}
\end{figure}

In the input layer of ECGFounder, the model receives a multi-channel ECG signal represented by a size of 12×5000, indicating 12 channels and 5000 time steps. This input passes through a 1D convolutional layer with an input size of 12×5000, where 64 filters of size 64×64 are applied to produce a feature map with dimensions of 64×160. As the network progresses beyond the first layer, multiple convolutional blocks are introduced, with each block composed of convolution and pooling operations that gradually increase the network depth. Initially, these transformations reduce the feature map size from 64×160 to 32×80, eventually leading to an output size of 8×400, achieved through two repetitions of the convolutional block.

Subsequently, from the third to the fifth layer, the network continues to apply convolution and pooling, which results in further dimensional reduction while progressively extracting more intricate features. The third layer's output remains at 400×400, with three repetitions of the convolution operation, while the fourth layer expands to produce a feature map of 1024×1024, repeated four times. Following these layers of convolution and feature extraction, the network transitions to a fully connected layer. Here, the extracted features from the preceding layers serve as input, and the fully connected operation generates either a single value or a categorical prediction based on the task at hand.

Notably, skip connections are included between layers to mitigate the vanishing gradient problem typically encountered in deep networks. These connections not only improve the training process but also enhance overall model performance by facilitating efficient information flow across the network.

\subsubsection*{S3.2. Details of training}

The training process for ECGFounder involved distinct sets of hyperparameters for pre-training and fine-tuning stages, optimized to achieve robust model performance. During the pre-training phase, a large batch size of 1024 was employed alongside a relatively high learning rate of 1e-3, which was gradually reduced using a ReduceLROnPlateau scheduler to a minimum of 1e-5. The AdamW optimizer was utilized with a weight decay of 0.1 to enhance generalization, and training was conducted over a total of 20 epochs.

In the fine-tuning phase, the batch size was reduced to 256, and the learning rate was lowered to 1e-4 with a minimum threshold of 1e-6. The ReduceLROnPlateau scheduler was retained to adaptively manage the learning rate, while the AdamW optimizer and a weight decay of 0.1 continued to be employed. The total number of epochs was increased to 30, allowing the model to refine its performance on downstream tasks. This systematic approach to hyperparameter selection ensured the model's adaptability and stability across various training stages.


\subsubsection*{S3.3. Details of ECG preprocessing}

Here, we provide a more comprehensive description of our preprocessing pipeline . The description is as follows:

{First, we used linear interpolation to resample each ECG signal to a fixed sampling rate of 500Hz. This step ensures that all ECG signals, regardless of their original sampling rate, have a consistent temporal resolution, facilitating downstream processing and analysis.
}

{Next, we applied a high-pass filter with a cutoff frequency of 1Hz to suppress residual baseline drift. Baseline drift often arises from low-frequency variations such as respiration and electrode-skin impedance changes. By attenuating frequency components below 1Hz, this high-pass filter helps maintain the main morphological features of the ECG waveform without low-frequency noise.
}

{After that, we employed a 2nd-order Butterworth low-pass filter with a cutoff frequency of 30Hz to reduce high-frequency noise. This filter preserves the primary frequency components of the ECG signal—typically within the 0–30Hz range—while attenuating higher frequencies that may be introduced by muscle artifacts or electrical interference.
}

{Furthermore, to remove power-line interference, we used a notch filter at 50/60Hz. This notch filter specifically targets and attenuates the narrow frequency band at which power-line noise occurs, improving signal fidelity without excessively affecting the rest of the ECG spectrum.
}

{Then, from any ECG recording longer than 10 seconds, we extracted consecutive 10-second segments. If a recording was shorter than 10 seconds, we applied zero-padding to the signal to reach the desired 10-second duration. This uniform window length ensures that each segment fed into the model has the same temporal dimension.
}

{Before inputting these processed segments into the model, we performed Z-score normalization using each segment’s mean and standard deviation. Concretely, this involves subtracting the segment mean from each data point and dividing by the segment’s standard deviation, resulting in each segment having a mean of 0 and a standard deviation of 1. This step prevents signals with inherently larger amplitudes from overshadowing those with smaller amplitudes, making model training more stable.
}

{Lastly, for any missing ECG signals, we simply assigned them a value of zero. In other words, these missing channels are represented as zero-filled data, ensuring a consistent input dimension for the model without introducing spurious numerical artifacts.
}

{By following these steps, we effectively mitigate baseline drift, high-frequency noise, and power-line interference, while establishing a uniform sampling rate and window size for all signals, ultimately providing clean and standardized ECG segments for subsequent analysis and model training.
}

\newpage

\subsection*{S4. Details of Experiments}

\subsubsection*{S4.1. Details of ablation experiments}

{We provide a series of ablation experimental results to demonstrate the advantages and innovations of our method.}

Here, we use the average AUC of the internal validation set and the average AUC of the CODE-test (external validation set) as our metrics, respectively.

The table are as follow:

\begin{table}[H]
\centering
\caption{Effects of loss function.}
\setlength{\tabcolsep}{10pt} %
\renewcommand{\arraystretch}{1.1} %
\label{tab:loss function}
\begin{tabular}{c|cc}
\toprule
\textbf{Loss} & \textbf{Internal} & \textbf{External}\\ 
\midrule
BCE Loss               & 86.18 (86.12-86.24)      & 95.32 (95.31-95.33)       \\
Focal Loss             & 88.67 (88.65-88.69)     & 96.73  (96.72-96.74)     \\
PU Loss (Ours)         & \textbf{92.04 (92.00-92.07)}     & \textbf{98.20  (98.15-98.25)}     \\

\bottomrule
\end{tabular}
\end{table}

\begin{table}[H]
\centering
\caption{Effects of hyperparameters of loss function.}
\setlength{\tabcolsep}{10pt} %
\renewcommand{\arraystretch}{1.1} %
\label{tab:loss function}
\begin{tabular}{c|cc}
\toprule
\textbf{Num of $gamma$} & \textbf{Internal} & \textbf{External}\\ 
\midrule
0.5               & 91.18 (91.14-91.22)      & 97.22 (97.21-97.23)       \\
1                & 91.77 (91.75-88.79)     & 97.37  (97.36-97.38)     \\
1.5 (Ours)         & \textbf{92.04 (92.00-92.07)}     & \textbf{98.20  (98.15-98.25)}     \\
2                & 91.96 (91.94-91.98)     & 98.12  (98.11-98.13)     \\

\bottomrule
\end{tabular}
\end{table}

\begin{table}[H]
\centering
\caption{Effects of model architecture.}
\setlength{\tabcolsep}{10pt} %
\renewcommand{\arraystretch}{1.1} %
\label{tab:model architecture}
\begin{tabular}{c|cc}
\toprule
\textbf{Model architecture} & \textbf{Internal} & \textbf{External}\\ 
\midrule
CNN-1d                     & 87.45 (87.43-97.47)     & 93.21 (93.12-93.30)       \\
ResNet-1d                 & 91.85  (91.84-91.86)    & 97.64 (97.62-97.67)       \\
Transformer-1d          & 91.53 (91.52-91.55)     & 97.78  (97.75-97.81)      \\
RegNet-1d (Ours)           & \textbf{92.04 (92.01-92.07)}     & \textbf{98.20 (98.18-98.23)}      \\

\bottomrule
\end{tabular}
\end{table}

\begin{table}[H]
\centering
\caption{Effects of 1-lead data argumentation.}
\setlength{\tabcolsep}{10pt} %
\renewcommand{\arraystretch}{1.1} %
\label{tab:1-lead data argumentation}
\begin{tabular}{c|cc}
\toprule
\textbf{Methods} & \textbf{Internal(1-lead)} & \textbf{External(1-lead)}\\ 
\midrule
Only 1-lead data                     & 83.76 (83.62-83.90)     & 94.96  (94.83-95.09)      \\

1-lead data argumentation (Ours)           & \textbf{86.92 (86.83-87.01)}     & \textbf{96.37 (96.33-96.41)}       \\

\bottomrule
\end{tabular}
\end{table}

\begin{table}[H]
\centering
\caption{Effects of number of model parameters.}
\setlength{\tabcolsep}{10pt} %
\renewcommand{\arraystretch}{1.1} %
\label{tab:model architecture}
\begin{tabular}{c|cc}
\toprule
\textbf{Model architecture} & \textbf{Internal} & \textbf{External}\\ 
\midrule
11.7M                     & 89.62 (89.50-89.74)     & 96.67 (96.61-96.73)       \\
25.6M                 & 90.23  (90.04-90.41)    & 97.41 (97.30-97.52)       \\
76.3M (Ours)          & \textbf{92.04 (92.01-92.07)}     & 98.20 (98.18-98.23)      \\
110M           &      91.87  (91.85-91.88)    & \textbf{98.34 (98.01-98.67)}  \\

\bottomrule
\end{tabular}
\end{table}

\begin{table}[H]
\centering
\caption{Effects of number of training set.}
\setlength{\tabcolsep}{10pt} %
\renewcommand{\arraystretch}{1.1} %
\label{tab:model architecture}
\begin{tabular}{c|cc}
\toprule
\textbf{Model architecture} & \textbf{Internal} & \textbf{External}\\ 
\midrule
10K                    & 86.78 (86.72-86.84)     & 91.11 (91.04-91.18)       \\
100K                    & 87.93 (87.93-87.93)     & 92.08 (92.03-92.13)       \\
1M                    & 89.95 (89.73-90.17)     & 96.21 (96.11-96.31)       \\
3M                & 91.85  (91.80-91.89)    & 97.82 (97.72-97.91)       \\
5M         & 91.93 (91.62-92.25)     & 97.78  (97.75-97.81)      \\
10M (Ours)           & \textbf{92.04 (92.01-92.07)}     & \textbf{98.20 (98.18-98.23)}      \\

\bottomrule
\end{tabular}
\end{table}

\newpage

\newpage

\subsubsection*{S4.2. Details of linear probing}

{In order to explore the advantages of ECGFounder under few-shot fine-tuning and the generalization of features, we conducted linear probing experiments on an ECG self supervised learning benchmark.}

{For linear probing, we kept the backbone network frozen and only trained the randomly initialized parameters of the linear classifier. To explore the performance of our method under low-resource conditions, we conducted linear probing using 1\%, 10\%, and 100\% of the training data for each task. We set the learning rate to $5\times10^{-3}$ and trained for 100 epochs.  All test results were obtained from the best validation model, rather than testing the model on the test set after each epoch and reporting the highest result. For all downstream tasks, we used the AUC as the evaluation metric. We set the random seed to $0$ to ensure the reproducibility of all results. 
}

\begin{table}
\caption{Linear probing results of ECGFounder and other SSL models. The best results are \textbf{bolded}, with \colorbox{gray!30}{gray} indicating the second highest.}
\label{tab:linear_prob}

\makebox[\linewidth]{ 
\resizebox{1\textwidth}{!}{
\setlength{\tabcolsep}{2pt} 
\renewcommand{\arraystretch}{1.6} 
\centering
\begin{tabular}{c|ccc|ccc|ccc|ccc|ccc|ccc} 
\toprule
\multicolumn{1}{c}{\multirow{2}{*}{\textbf{Method}}} & \multicolumn{3}{c}{\textbf{PTBXL-Super}}                                                 & \multicolumn{3}{c}{\textbf{PTBXL-Sub}}                                                   & \multicolumn{3}{c}{\textbf{PTBXL-Form}}                                                  & \multicolumn{3}{c}{\textbf{PTBXL-Rhythm}}                                                & \multicolumn{3}{c}{\textbf{CPSC2018}}                                                    & \multicolumn{3}{c}{\textbf{CSN}}                                                         \\
\multicolumn{1}{c}{}                        & \multicolumn{1}{c}{\textbf{1\%}} & \multicolumn{1}{c}{\textbf{10\%}} & \multicolumn{1}{c|}{\textbf{100\%}} & \multicolumn{1}{c}{\textbf{1\%}} & \multicolumn{1}{c}{\textbf{10\%}} & \multicolumn{1}{c|}{\textbf{100\%}} & \multicolumn{1}{c}{\textbf{1\%}} & \multicolumn{1}{c}{\textbf{10\%}} & \multicolumn{1}{c|}{\textbf{100\%}} & \multicolumn{1}{c}{\textbf{1\%}} & \multicolumn{1}{c}{\textbf{10\%}} & \multicolumn{1}{c|}{\textbf{100\%}} & \multicolumn{1}{c}{\textbf{1\%}} & \multicolumn{1}{c}{\textbf{10\%}} & \multicolumn{1}{c|}{\textbf{100\%}} & \multicolumn{1}{c}{\textbf{1\%}} & \multicolumn{1}{c}{\textbf{10\%}} & \multicolumn{1}{c}{\textbf{100\%}}  \\ 

\midrule

SimCLR \citep{pmlr-v119-chen20j}                                   & 63.41                   & 69.77                    & 73.53                      & 60.84                   & 68.27                    & 73.39                      & 54.98                   & 56.97                    & 62.52                      & 51.41                   & 69.44                    & 77.73                      & 59.78                   & 68.52                    & 76.54                      & 59.02                   & 67.26                    & 73.20                      \\

TS-TCC \citep{eldele2021time}                                     & 70.73                   & 75.88                    & 78.91                      & 53.54                   & 66.98                    & 77.87                      & 48.04                   & 61.79                    & 71.18                      & 43.34                   & 69.48                    & 78.23                      & 57.07                   & 73.62                    & 78.72                      & 55.26                   & 68.48                    & 76.79                      \\
CLOCS \citep{kiyasseh2021clocs}                                      & 68.94                   & 73.36                    & 76.31                      & 57.94                   & 72.55                   & 76.24                      & 51.97                   & 57.96                    & 72.65                      & 47.19                   & 71.88                    & 76.31                      & 59.59                   & \colorbox{gray!30}{77.78}                    & 77.49                      & 54.38                   & \colorbox{gray!30}{71.93}                    & 76.13                      \\

ST-MEM \citep{na2024guiding}                                     & 61.12                   & 66.87                    & 71.36                      & 54.12                   & 57.86                    & 63.59                      & 55.71                  & 59.99                    & 66.07                      & 51.12                   & 65.44                    & 74.85                      & 56.69                   & 63.32                    & 70.39                      & \colorbox{gray!30}{59.77}                   & 66.87                    & 71.36                      \\ 

HeartLang \citep{jinreading}                              & \colorbox{gray!30}{78.94}                  & \colorbox{gray!30}{85.59}                   & \colorbox{gray!30}{87.52}                   & \colorbox{gray!30}{64.68}                 & \colorbox{gray!30}{79.34}                 & \textbf{88.91}                      & 58.70                   & 63.99                    & \colorbox{gray!30}{80.23}                     & \colorbox{gray!30}{62.08}                 & \colorbox{gray!30}{76.22}                    & \colorbox{gray!30}{90.34}                     & \colorbox{gray!30}{60.44}                   & 66.26                    & 77.87                      & 57.94                   & 68.93                    & \colorbox{gray!30}{82.49}                      \\

\midrule
\textbf{ECGFounder (Ours) }                            & \textbf{79.65}                  & \textbf{87.34}                    & \textbf{91.20}                    & \textbf{75.72}                   & \textbf{81.73}                    & \colorbox{gray!30}{88.03}                      & \textbf{61.69}                   & \textbf{68.76}                    & \textbf{82.52}                      & \textbf{70.05}                  & \textbf{88.10}                    & \textbf{93.61}                     & \textbf{64.21}                   & \textbf{79.65}                    & \textbf{83.18}                     & \textbf{64.68}                   & \textbf{75.89}                    & \textbf{84.35}                      \\
\bottomrule
\end{tabular}
}
}
\end{table}

\newpage

\subsubsection*{S4.3. Details of subgroup analysis}

{In order to explore the fairness of ECGFounder under various demographic conditions, we conducted subgroup analysis on ECGFounder. The table is as follows: }

\begin{table}[H]
\centering
\caption{Performance in age, sex, and race stratified populations.}
\setlength{\tabcolsep}{10pt} %
\renewcommand{\arraystretch}{1.1} %
\label{tab:model architecture}
\begin{tabular}{c|cc}
\toprule
\textbf{Cohort} & \textbf{Internal} & \textbf{External}\\ 
\midrule
Less than 40 years old       & 91.85 (91.75-91.94)     & 98.21 (98.18-98.24)       \\
40-60 years old              & 92.44 (92.39-92.48)     & 97.94 (97.92-97.96)       \\
60-75 years old              & 91.58 (91.55-91.62)     & 97.78 (97.75-97.81)       \\
Greater than 75 years old    & 92.03 (91.95-92.13)      & 98.21 (98.20-98.23)       \\
Male                         & 91.98 (91.92-92.14)      & 98.22 (98.18-98.26)       \\
Female                       & 92.10 (92.06-92.15)     & 98.16 (98.15-98.17)       \\
White                        & 92.72 (92.66-92.78)     & /        \\
Asian                        & 91.44 (91.40-92.48)     & /        \\
African American             & 91.90 (91.86-91.94)     & /       \\
American Indian              & 90.55 (90.46-90.64)     & /       \\
Other races                  & 92.11 (92.07-92.15)     & /        \\
\bottomrule
\end{tabular}
\end{table}

\newpage

\subsection*{S5. Details of Results}

\subsubsection*{S5.1. Details of internal committee validation}

\begin{figure}[h]
    \centering
    \begin{adjustbox}{width=\linewidth,center}
    \begin{tabular}{cccc}
        \begin{subfigure}[b]{0.25\textwidth}
            \centering
            \includegraphics[width=\textwidth]{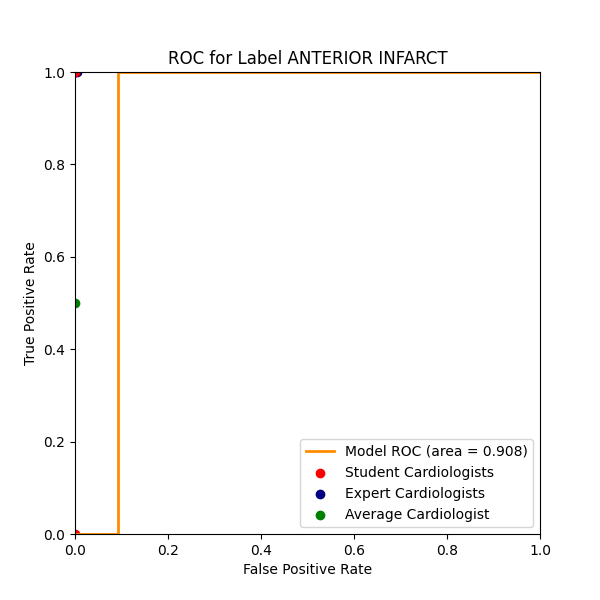}
        \end{subfigure} &
        \begin{subfigure}[b]{0.25\textwidth}
            \centering
            \includegraphics[width=\textwidth]{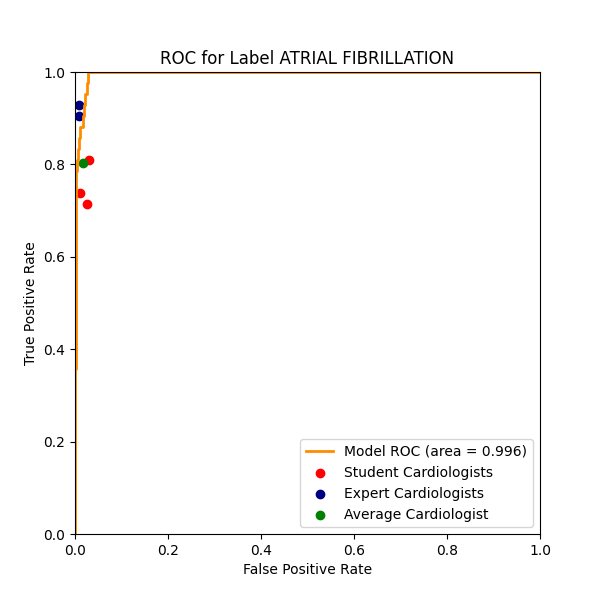}
        \end{subfigure} &
        \begin{subfigure}[b]{0.25\textwidth}
            \centering
            \includegraphics[width=\textwidth]{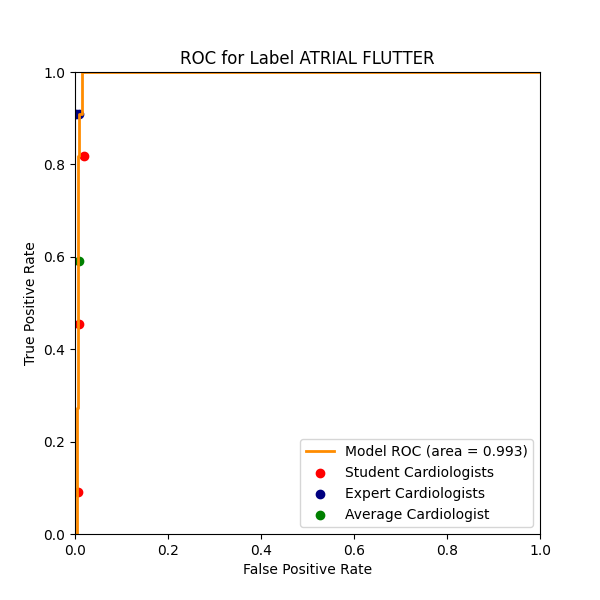}
        \end{subfigure} &
        \begin{subfigure}[b]{0.25\textwidth}
            \centering
            \includegraphics[width=\textwidth]{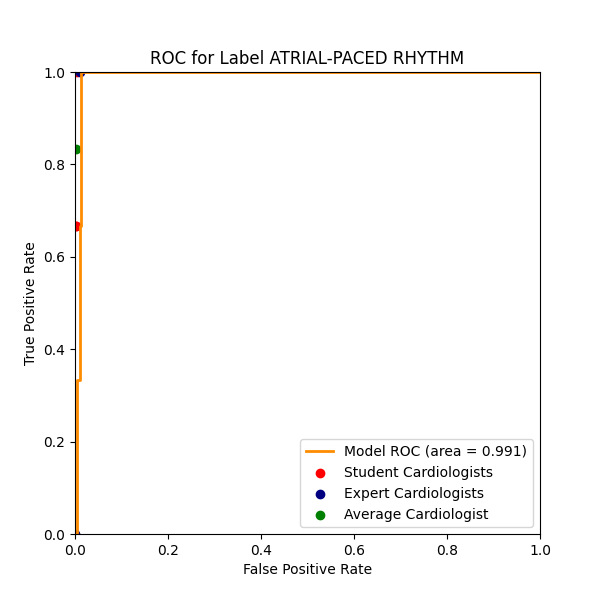}
        \end{subfigure} \\
        \begin{subfigure}[b]{0.25\textwidth}
            \centering
            \includegraphics[width=\textwidth]{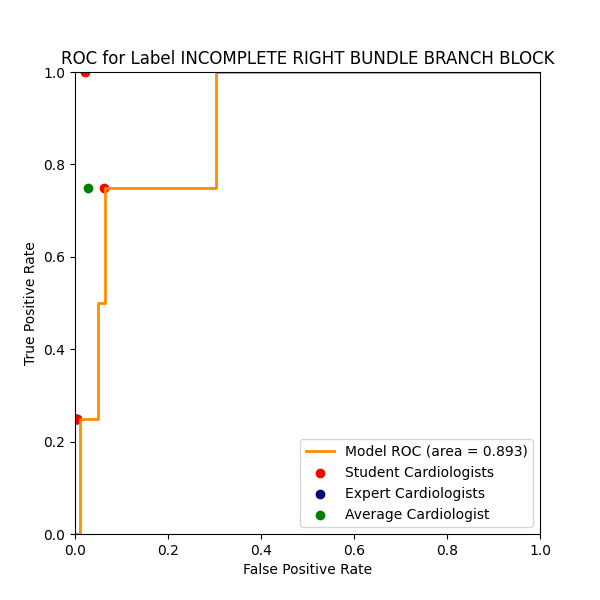}
        \end{subfigure} &
        \begin{subfigure}[b]{0.25\textwidth}
            \centering
            \includegraphics[width=\textwidth]{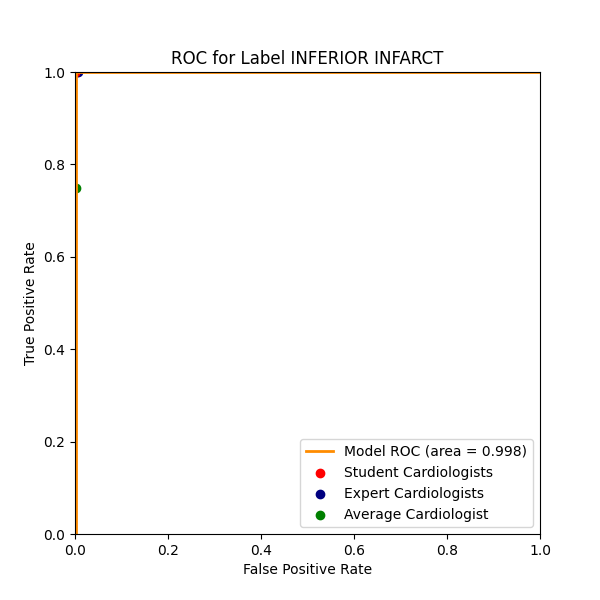}
        \end{subfigure} &
        \begin{subfigure}[b]{0.25\textwidth}
            \centering
            \includegraphics[width=\textwidth]{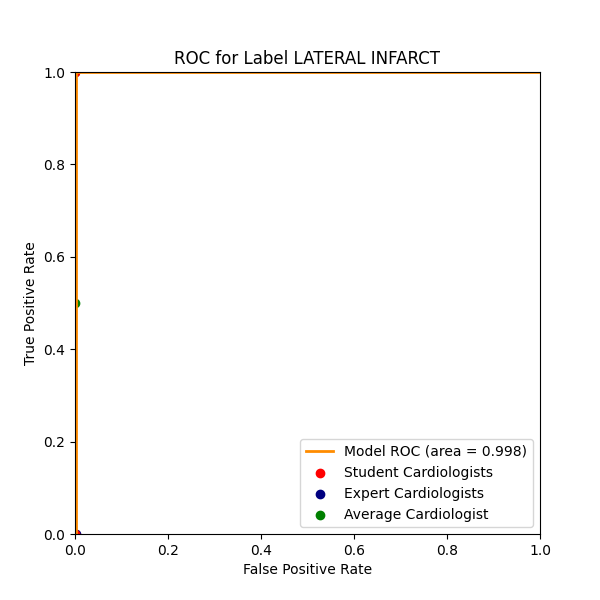}
        \end{subfigure} &
        \begin{subfigure}[b]{0.25\textwidth}
            \centering
            \includegraphics[width=\textwidth]{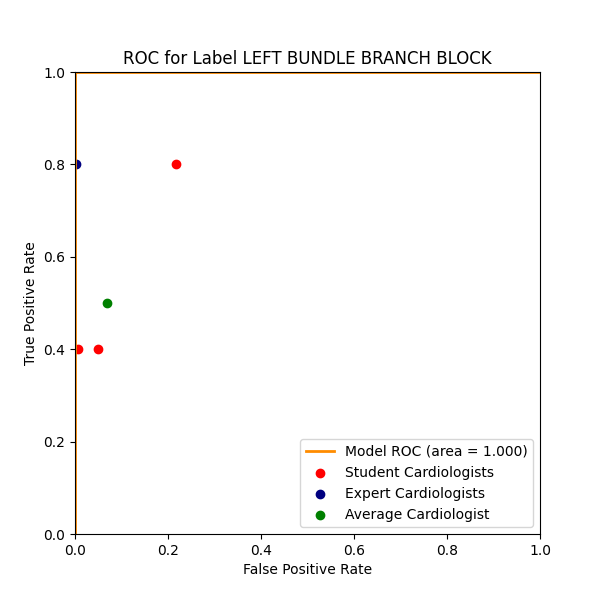}
        \end{subfigure} \\
        \begin{subfigure}[b]{0.25\textwidth}
            \centering
            \includegraphics[width=\textwidth]{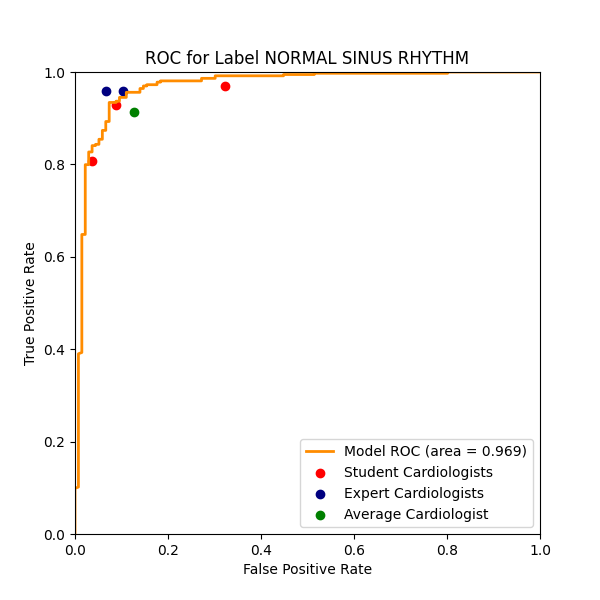}
        \end{subfigure} &
        \begin{subfigure}[b]{0.25\textwidth}
            \centering
            \includegraphics[width=\textwidth]{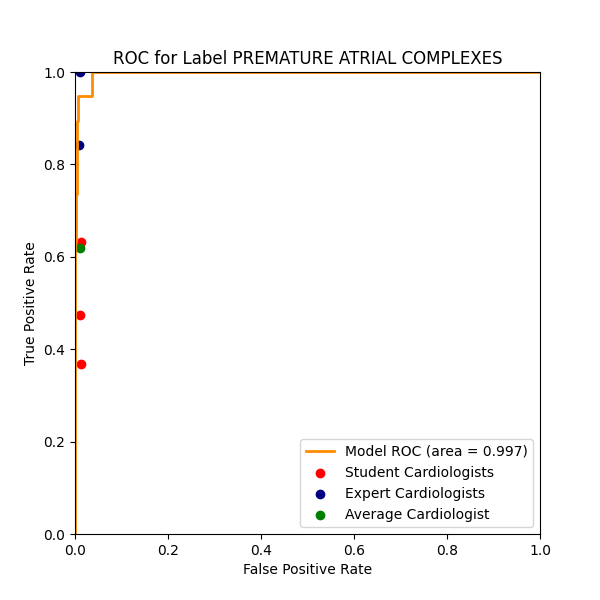}
        \end{subfigure} &
        \begin{subfigure}[b]{0.25\textwidth}
            \centering
            \includegraphics[width=\textwidth]{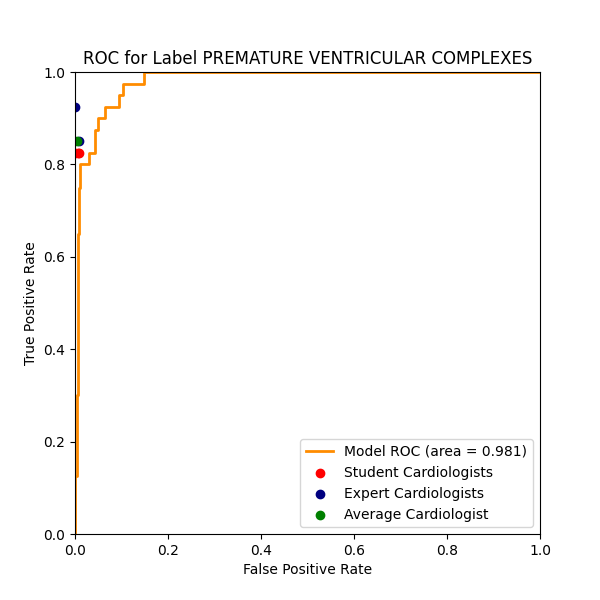}
        \end{subfigure} &
        \begin{subfigure}[b]{0.25\textwidth}
            \centering
            \includegraphics[width=\textwidth]{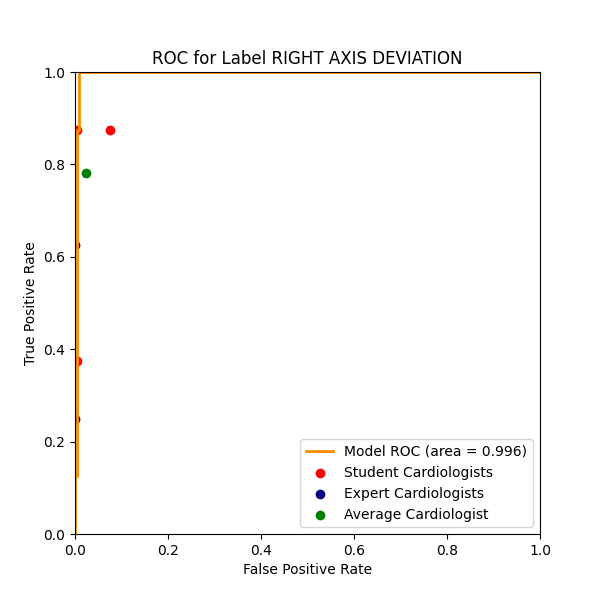}
        \end{subfigure} \\
        \begin{subfigure}[b]{0.25\textwidth}
            \centering
            \includegraphics[width=\textwidth]{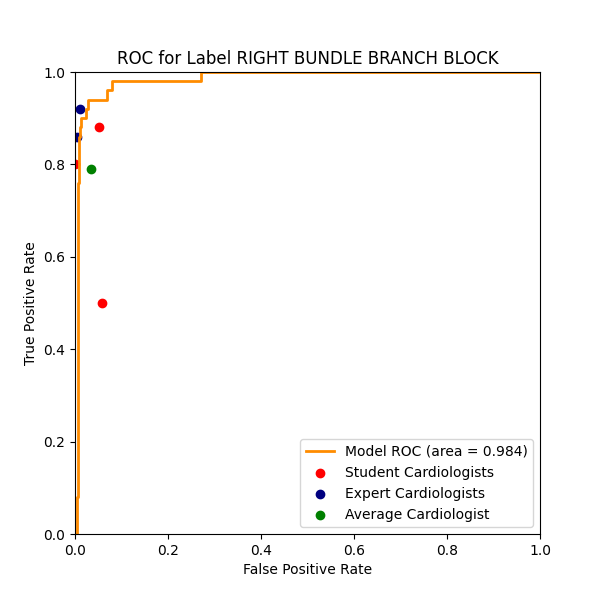}
        \end{subfigure} &
        \begin{subfigure}[b]{0.25\textwidth}
            \centering
            \includegraphics[width=\textwidth]{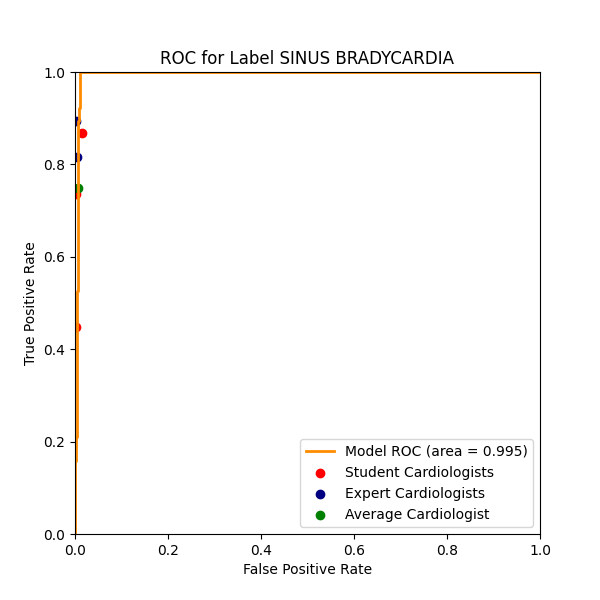}
        \end{subfigure} &
        \begin{subfigure}[b]{0.25\textwidth}
            \centering
            \includegraphics[width=\textwidth]{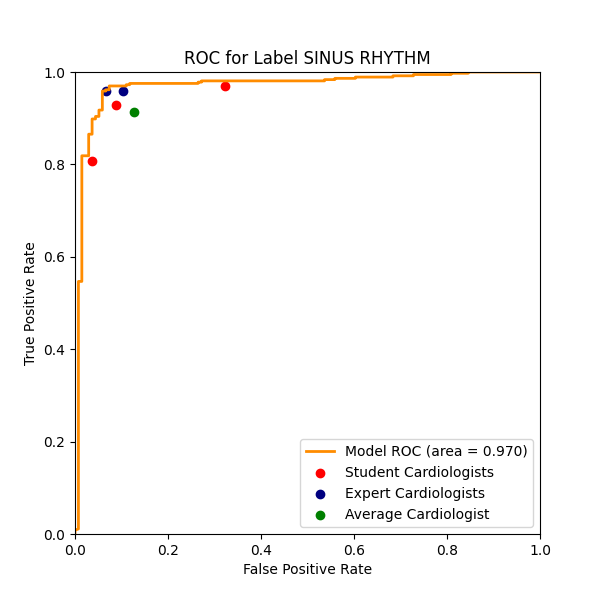}
        \end{subfigure} &
        \begin{subfigure}[b]{0.25\textwidth}
            \centering
            \includegraphics[width=\textwidth]{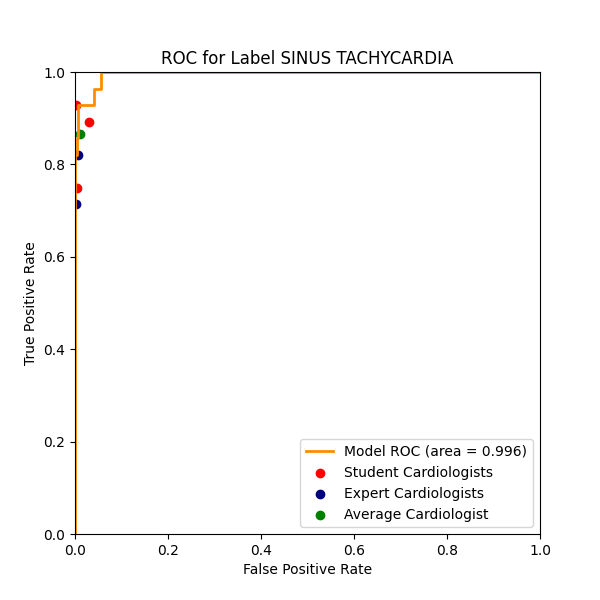}
        \end{subfigure} \\
        \begin{subfigure}[b]{0.25\textwidth}
            \centering
            \includegraphics[width=\textwidth]{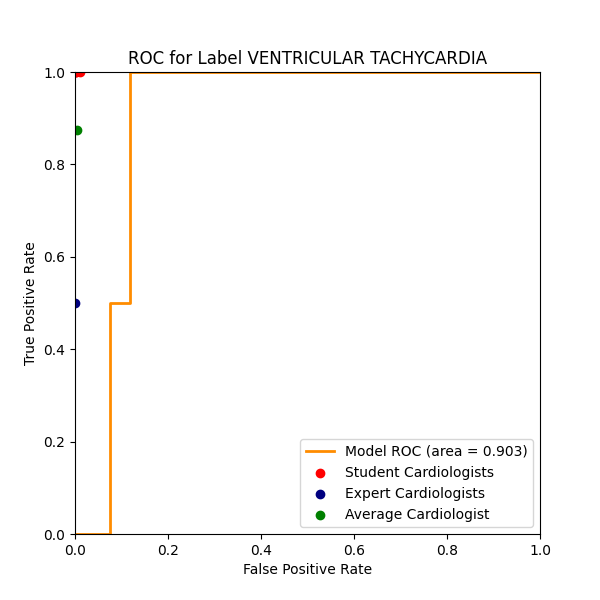}
        \end{subfigure} &
        \begin{subfigure}[b]{0.25\textwidth}
            \centering
            \includegraphics[width=\textwidth]{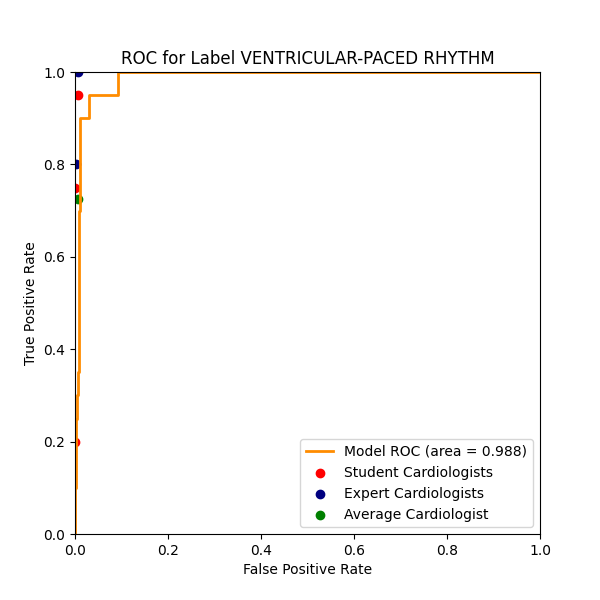}
        \end{subfigure} &
        \begin{subfigure}[b]{0.25\textwidth}
            \centering
            \includegraphics[width=\textwidth]{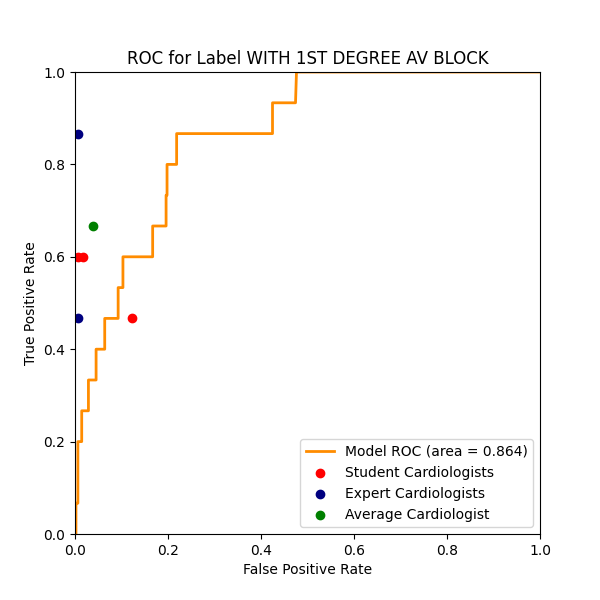}
        \end{subfigure} &
        \begin{subfigure}[b]{0.25\textwidth}
            \centering
            \includegraphics[width=\textwidth]{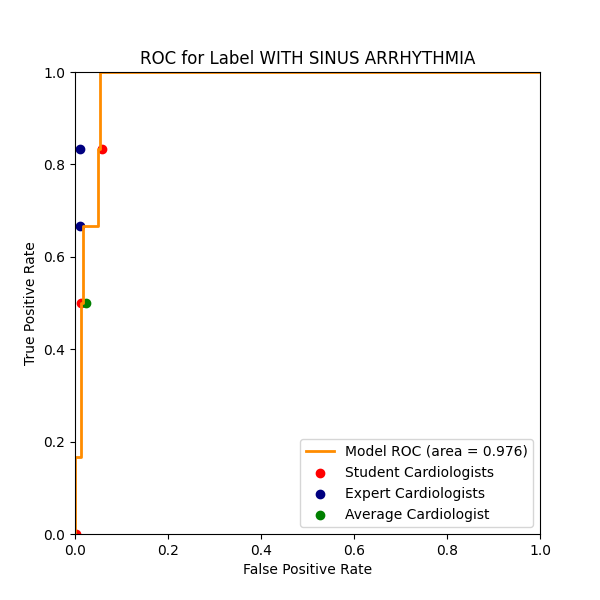}
        \end{subfigure} \\
    \end{tabular}
    \end{adjustbox}
    \caption{ROC curves on the internal test set of the deep learning model, compared with cardiologists}
    \label{fig:committee}
\end{figure}

To further compare the performance of ECGFounder in clinical diagnosis with that of experts, we compared the performance of the model with that of 5 cardiologists. The results are shown in the figure \ref{fig:committee}.

\begin{table}[]
\begin{tabular}{|l|l|l|l|l|}
\hline
Label                                & Sensitivity & Sensitivity\_CI & Specificity & Specificity\_CI \\ \hline
LATERAL INFARCT                      & 1           & (1.0, 1.0)      & 0.94        & (0.92, 0.96)    \\ \hline
INFERIOR INFARCT                     & 1           & (1.0, 1.0)      & 1           & (0.99, 0.99)    \\ \hline
PREMATURE ATRIAL COMPLEXES           & 0.95        & (0.87, 1.0)     & 0.96        & (0.95, 0.97)    \\ \hline
SINUS TACHYCARDIA                    & 0.93        & (0.85, 1.0)     & 0.97        & (0.96, 0.99)    \\ \hline
RIGHT AXIS DEVIATION                 & 1           & (1.0, 1.0)      & 0.97        & (0.96, 0.98)    \\ \hline
ATRIAL FIBRILLATION                  & 1           & (1.0, 1.0)      & 0.97        & (0.95, 0.98)    \\ \hline
SINUS BRADYCARDIA                    & 1           & (1.0, 1.0)      & 0.98        & (0.97, 0.99)    \\ \hline
ATRIAL FLUTTER                       & 0.91        & (0.90, 0.94)    & 0.99        & (0.98, 0.99)    \\ \hline
ATRIAL-PACED RHYTHM                  & 0.63        & (0.32, 0.94)    & 1           & (0.99, 1.0)     \\ \hline
RIGHT BUNDLE BRANCH BLOCK            & 0.92        & (0.88, 0.97)    & 0.97        & (0.95, 0.98)    \\ \hline
PREMATURE VENTRICULAR COMPLEXES      & 0.88        & (0.79, 0.97)    & 0.95        & (0.93, 0.97)    \\ \hline
WITH SINUS ARRHYTHMIA                & 0.75        & (0.67, 0.87)    & 0.96        & (0.94, 0.96)    \\ \hline
SINUS RHYTHM                         & 0.98        & (0.96, 0.99)    & 0.54        & (0.48, 0.63)    \\ \hline
NORMAL SINUS RHYTHM                  & 0.94        & (0.91, 0.96)    & 0.9         & (0.89, 0.95)    \\ \hline
ANTERIOR INFARCT                     & 0           & (0.0, 0.0)      & 0.93        & (0.91, 0.95)    \\ \hline
VENTRICULAR TACHYCARDIA              & 0           & (0.0, 0.0)      & 1           & (0.99, 1.0)     \\ \hline
INCOMPLETE RIGHT BUNDLE BRANCH BLOCK & 0.93        & (0.85, 1.0)     & 0.99        & (0.98, 1.0)     \\ \hline
WITH 1ST DEGREE AV BLOCK             & 0.66        & (0.62, 0.68)    & 1           & (0.99, 1.0)     \\ \hline
LEFT BUNDLE BRANCH BLOCK             & 1           & (1.0, 1.0)      & 0.98        & (0.97, 0.99)    \\ \hline
VENTRICULAR-PACED RHYTHM             & 0.94        & (0.91, 0.96)    & 1           & (0.99, 1.0)     \\ \hline
\end{tabular}
\end{table}

\begin{table}[]
\begin{tabular}{|l|l|l|l|l|}
\hline
Label                                & PPV  & PPV\_CI      & NPV  & NPV\_CI      \\ \hline
LATERAL INFARCT                      & 0.03 & (0.0, 0.10)  & 1    & (1.0, 1.0)   \\ \hline
INFERIOR INFARCT                     & 0.33 & (0.0, 0.66)  & 1    & (1.0, 1.0)   \\ \hline
PREMATURE ATRIAL COMPLEXES           & 0.5  & (0.39, 0.62) & 1    & (1.0, 1.0)   \\ \hline
SINUS TACHYCARDIA                    & 0.68 & (0.51, 0.81) & 1    & (0.99, 1.0)  \\ \hline
RIGHT AXIS DEVIATION                 & 0.35 & (0.27, 0.49) & 1    & (1.0, 1.0)   \\ \hline
ATRIAL FIBRILLATION                  & 0.75 & (0.67, 0.87) & 1    & (1.0, 1.0)   \\ \hline
SINUS BRADYCARDIA                    & 0.81 & (0.67, 0.89) & 1    & (1.0, 1.0)   \\ \hline
ATRIAL FLUTTER                       & 0.63 & (0.32, 0.94) & 1    & (1.0, 1.0)   \\ \hline
ATRIAL-PACED RHYTHM                  & 0.68 & (0.51, 0.81) & 1    & (0.99, 1.0)  \\ \hline
RIGHT BUNDLE BRANCH BLOCK            & 0.78 & (0.70, 0.88) & 0.99 & (0.98, 1.0)  \\ \hline
PREMATURE VENTRICULAR COMPLEXES      & 0.63 & (0.54, 0.74) & 0.99 & (0.98, 1.0)  \\ \hline
WITH SINUS ARRHYTHMIA                & 0.15 & (0.05, 0.25) & 1    & (0.99, 1.0)  \\ \hline
SINUS RHYTHM                         & 0.85 & (0.81, 0.88) & 0.91 & (0.86, 0.98) \\ \hline
NORMAL SINUS RHYTHM                  & 0.96 & (0.95, 0.97) & 0.84 & (0.80, 0.92) \\ \hline
ANTERIOR INFARCT                     & 0    & (0.0, 0.0)   & 1    & (1.0, 1.0)   \\ \hline
VENTRICULAR TACHYCARDIA              & 0    & (0.0, 0.0)   & 1    & (0.99, 1.0)  \\ \hline
INCOMPLETE RIGHT BUNDLE BRANCH BLOCK & 0.13 & (0.0, 0.34)  & 0.99 & (0.99, 1.0)  \\ \hline
WITH 1ST DEGREE AV BLOCK             & 0.5  & (0.0, 1.0)   & 0.97 & (0.96, 0.99) \\ \hline
LEFT BUNDLE BRANCH BLOCK             & 0.33 & (0.10, 0.51) & 1    & (1.0, 1.0)   \\ \hline
VENTRICULAR-PACED RHYTHM             & 0.71 & (0.71, 1.0)  & 0.97 & (0.96, 0.98) \\ \hline
\end{tabular}
\end{table}

\begin{table}[]
\begin{tabular}{|l|l|l|l|l|}
\hline
Label                                & AUROC & AUROC\_CI    & AUPRC & AUPRC\_CI    \\ \hline
LATERAL INFARCT                      & 1     & (1.0, 1.0)   & 0.62  & (0.57, 0.67) \\ \hline
INFERIOR INFARCT                     & 1     & (1.0, 1.0)   & 0.5   & (0.39, 0.61) \\ \hline
PREMATURE ATRIAL COMPLEXES           & 1     & (0.99, 1.0)  & 0.94  & (0.86, 0.98) \\ \hline
SINUS TACHYCARDIA                    & 1     & (0.99, 1.0)  & 0.96  & (0.90, 0.99) \\ \hline
RIGHT AXIS DEVIATION                 & 1     & (0.99, 1.0)  & 0.71  & (0.29, 0.95) \\ \hline
ATRIAL FIBRILLATION                  & 1     & (0.99, 1.0)  & 0.95  & (0.87, 0.98) \\ \hline
SINUS BRADYCARDIA                    & 1     & (0.99, 1.0)  & 0.9   & (0.83, 0.99) \\ \hline
ATRIAL FLUTTER                       & 0.99  & (0.99, 1.0)  & 0.62  & (0.34, 0.94) \\ \hline
ATRIAL-PACED RHYTHM                  & 0.99  & (0.99, 1.0)  & 0.35  & (0.27, 0.49) \\ \hline
RIGHT BUNDLE BRANCH BLOCK            & 0.98  & (0.97, 0.99) & 0.81  & (0.65, 0.89) \\ \hline
PREMATURE VENTRICULAR COMPLEXES      & 0.98  & (0.97, 0.99) & 0.81  & (0.71, 0.91) \\ \hline
WITH SINUS ARRHYTHMIA                & 0.98  & (0.96, 1.0)  & 0.37  & (0.11, 0.49) \\ \hline
SINUS RHYTHM                         & 0.97  & (0.96, 0.99) & 0.98  & (0.95, 1.0)  \\ \hline
NORMAL SINUS RHYTHM                  & 0.97  & (0.94, 0.98) & 0.98  & (0.97, 0.99) \\ \hline
ANTERIOR INFARCT                     & 0.91  & (0.86, 0.98) & 0.66  & (0.62, 0.68) \\ \hline
VENTRICULAR TACHYCARDIA              & 0.9   & (0.89, 0.95) & 0.54  & (0.48, 0.63) \\ \hline
INCOMPLETE RIGHT BUNDLE BRANCH BLOCK & 0.88  & (0.79, 0.97) & 0.33  & (0.10, 0.51) \\ \hline
WITH 1ST DEGREE AV BLOCK             & 0.86  & (0.80, 0.92) & 0.22  & (0.13, 0.41) \\ \hline
LEFT BUNDLE BRANCH BLOCK             & 1     & (1.0, 1.0)   & 1     & (1.0, 1.0)   \\ \hline
VENTRICULAR-PACED RHYTHM             & 0.99  & (0.98, 1.0)  & 0.74  & (0.58, 0.89) \\ \hline
\end{tabular}
\end{table}

\begin{table}[]
\centering
\caption{The evaluation results of each cardiologist.}
\begin{tabular}{l|ccc|ccc}
\toprule
\label{tab:internal}
{}  & \multicolumn{3}{c}{Cardiologist 1} & \multicolumn{3}{c}{Cardiologist 2} \\ \midrule
 
Class                                & Sens &  Spec & F1 & Sens &  Spec & F1\\
\midrule
ANTERIOR INFARCT                     & 1.000       & 1.000       & 1.000 & 1.000       & 0.996       & 0.500 \\
ATRIAL FIBRILLATION                  & 0.810       & 0.969       & 0.756 & 0.929       & 0.991       & 0.918 \\
ATRIAL FLUTTER                       & 0.818       & 0.982       & 0.521 & 0.909       & 0.992       & 0.544 \\
ATRIAL-PACED RHYTHM                  & 1.000       & 0.996       & 0.750 & 1.000       & 0.990       & 0.545 \\
INCOMPLETE RIGHT BUNDLE BRANCH BLOCK & 0.750       & 0.938       & 0.248 & 0.250       & 0.998       & 0.333 \\
INFERIOR INFARCT                     & 1.000       & 0.996       & 0.500 & 1.000       & 0.994       & 0.400 \\
LATERAL INFARCT                      & 0.000       & 1.000       & 0.000 & 0.000       & 0.998       & 0.000 \\
LEFT BUNDLE BRANCH BLOCK             & 0.454       & 0.949       & 0.125 & 0.727       & 1.000       & 0.189 \\
NORMAL SINUS RHYTHM                  & 0.929       & 0.912       & 0.947 & 0.959       & 0.897       & 0.960 \\
PREMATURE ATRIAL COMPLEXES           & 0.368       & 0.988       & 0.438 & 0.842       & 0.992       & 0.821 \\
PREMATURE VENTRICULAR COMPLEXES      & 0.830       & 0.993       & 0.868 & 0.930       & 1.000       & 0.961 \\
RIGHT AXIS DEVIATION                 & 0.380       & 0.996       & 0.482 & 0.250       & 1.000       & 0.498 \\
RIGHT BUNDLE BRANCH BLOCK            & 0.500       & 0.942       & 0.495 & 0.920       & 0.989       & 0.911 \\
SINUS BRADYCARDIA                    & 0.737       & 0.998       & 0.736 & 0.816       & 0.996       & 0.873 \\
SINUS RHYTHM                         & 0.929       & 0.912       & 0.947 & 0.959       & 0.897       & 0.960 \\
SINUS TACHYCARDIA                    & 0.929       & 0.998       & 0.945 & 0.714       & 0.998       & 0.816 \\
VENTRICULAR TACHYCARDIA              & 1.000       & 0.998       & 0.667 & 0.500       & 1.000       & 0.566 \\
VENTRICULAR-PACED RHYTHM             & 0.950       & 0.994       & 0.705 & 1.000       & 0.994       & 0.730 \\
WITH 1ST DEGREE AV BLOCK             & 0.600       & 0.994       & 0.667 & 0.867       & 0.994       & 0.639 \\
WITH SINUS ARRHYTHMIA                & 0.000       & 0.998       & 0.167 & 0.833       & 0.990       & 0.517 \\
\bottomrule

\end{tabular}
\end{table}

\begin{table}[]
\centering
\caption{The evaluation results of each cardiologist.}
\begin{tabular}{l|ccc|ccc}
\toprule
\label{tab:internal}
{}  & \multicolumn{3}{c}{Cardiologist 3} & \multicolumn{3}{c}{Cardiologist 4} \\ \midrule
 
Class                                & Sens &  Spec & F1 & Sens &  Spec & F1\\
\midrule

ANTERIOR INFARCT                     & 1.000       & 1.000       & 1.000 & 0.000       & 1.000       & 0.000 \\
ATRIAL FIBRILLATION                  & 0.738       & 0.991       & 0.805 & 0.714       & 0.974       & 0.714 \\
ATRIAL FLUTTER                       & 0.091       & 0.998       & 0.409 & 0.455       & 0.992       & 0.493 \\
ATRIAL-PACED RHYTHM                  & 0.667       & 1.000       & 0.657 & 0.667       & 0.998       & 0.667 \\
INCOMPLETE RIGHT BUNDLE BRANCH BLOCK & 1.000       & 0.996       & 0.286 & 0.250       & 0.996       & 0.286 \\
INFERIOR INFARCT                     & 1.000       & 1.000       & 1.000 & 1.000       & 1.000       & 1.000 \\
LATERAL INFARCT                      & 0.000       & 1.000       & 0.000 & 1.000       & 1.000       & 1.000 \\
LEFT BUNDLE BRANCH BLOCK             & 0.727       & 0.998       & 0.800 & 0.454       & 0.994       & 0.400 \\
NORMAL SINUS RHYTHM                  & 0.970       & 0.934       & 0.967 & 0.808       & 0.963       & 0.887 \\
PREMATURE ATRIAL COMPLEXES           & 0.474       & 0.990       & 0.684 & 0.632       & 0.988       & 0.649 \\
PREMATURE VENTRICULAR COMPLEXES      & 0.830       & 0.991       & 0.872 & 0.830       & 1.000       & 0.904 \\
RIGHT AXIS DEVIATION                 & 0.880       & 1.000       & 0.669 & 0.880       & 0.996       & 0.534 \\
RIGHT BUNDLE BRANCH BLOCK            & 0.880       & 0.996       & 0.805 & 0.800       & 1.000       & 0.889 \\
SINUS BRADYCARDIA                    & 0.447       & 0.998       & 0.632 & 0.868       & 0.985       & 0.886 \\
SINUS RHYTHM                         & 0.970       & 0.934       & 0.967 & 0.808       & 0.963       & 0.887 \\
SINUS TACHYCARDIA                    & 0.750       & 0.994       & 0.852 & 0.893       & 0.970       & 0.746 \\
VENTRICULAR TACHYCARDIA              & 1.000       & 1.000       & 0.800 & 1.000       & 0.998       & 0.667 \\
VENTRICULAR-PACED RHYTHM             & 0.200       & 0.998       & 0.885 & 0.750       & 1.000       & 0.657 \\
WITH 1ST DEGREE AV BLOCK             & 0.467       & 0.994       & 0.560 & 0.600       & 0.984       & 0.563 \\
WITH SINUS ARRHYTHMIA                & 0.500       & 0.990       & 0.311 & 0.833       & 0.941       & 0.250 \\
\bottomrule
\end{tabular}
\end{table}

newpage

\subsubsection*{S5.2. Details of internal validation}

\begin{figure}[hbp]
    \centering
    \includegraphics[width=\textwidth]{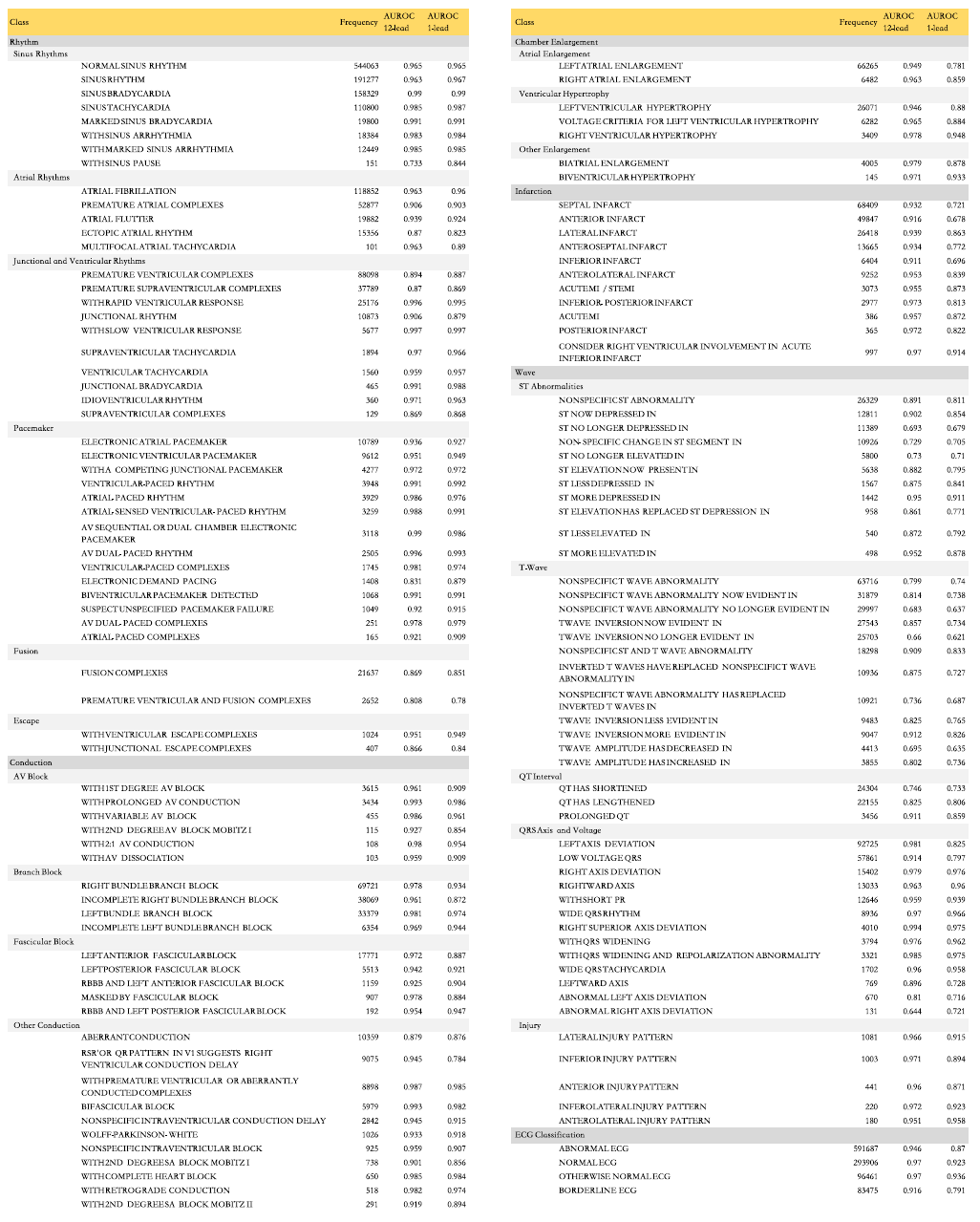} 
    \caption{Results of AUROC on 12-lead and single-lead ECG in internal test set.}
    \label{fig:eval}
\end{figure}

\newpage

\subsubsection*{S5.3. Details of external validation}

We evaluated single-lead atrial fibrillation detection on an external validation set. As shown in the table\ref{tab:2017}, the performance of our model exceeds the previous SOTA method. 

\begin{table*}[h]
 \centering
\caption{Performance of other ECG deep learning model and ECGFounder on external test set (single-lead ECG) PhysioNet Challenge-2017}
\label{tab:2017}
\begin{tabular}{l|cccc|c}
\toprule
{Models} & \multicolumn{4}{c}{ECGFounder} & \multicolumn{1}{c}{Stanford\cite{hannunCardiologistlevelArrhythmiaDetection2019}} \\\midrule

Class & AUC   & Sens & Spec & F1 & F1     \\ 
\midrule

SINUS RHYTHM & 0.975 & 0.961   & 0.983     & \textbf{0.941}  & 0.910\\
ATRIAL FIBRILLATION & 0.957 & 0.924   & 0.979     &  \textbf{0.856} & 0.840\\
\bottomrule
\end{tabular}
\end{table*}

\newpage

\subsubsection*{S5.4. Details of fine-tuning}

\begin{figure}[h]
    \centering
    \includegraphics[width=\textwidth]{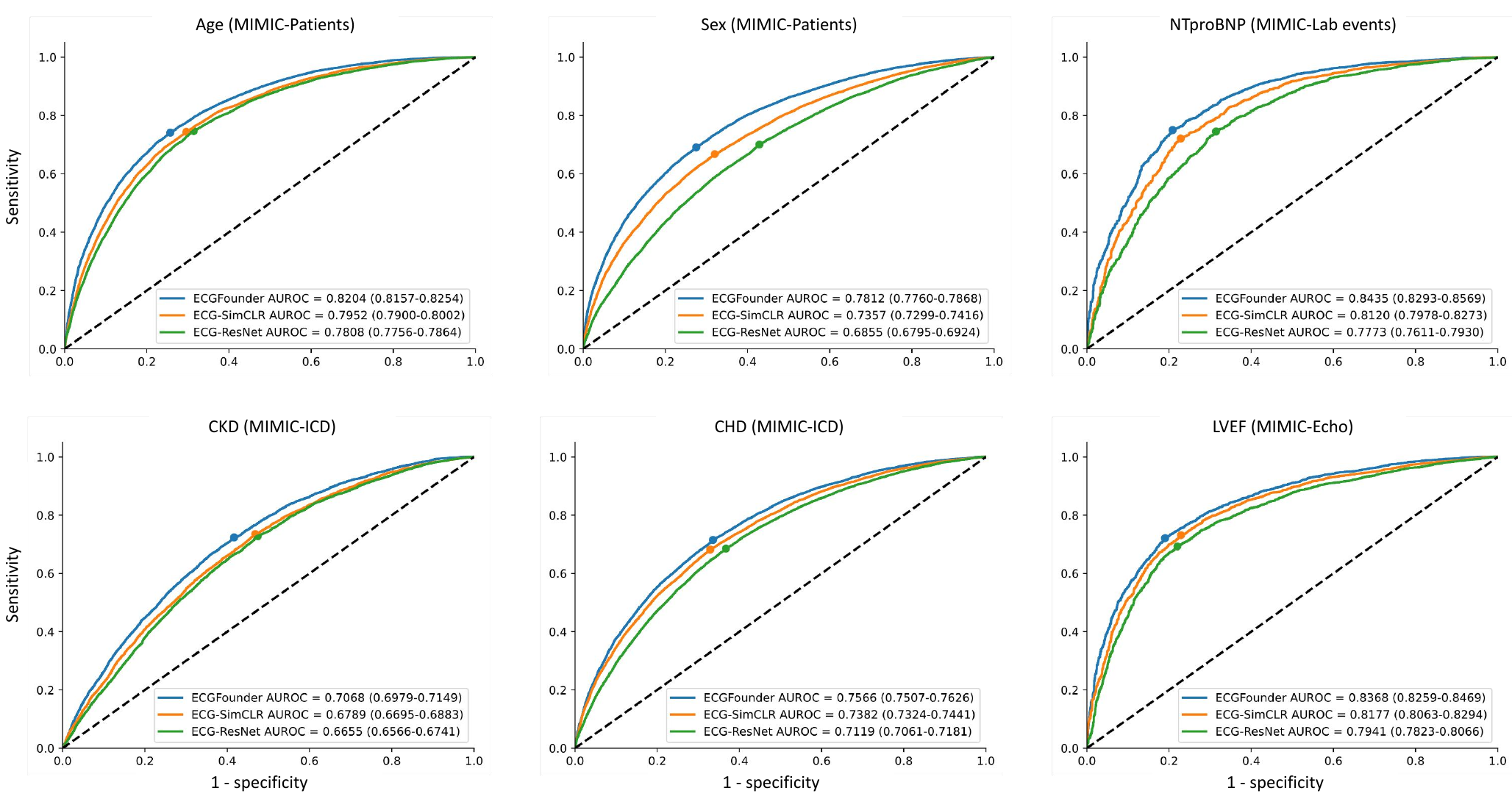} 
    \caption{AUROC of single-lead ECG downstream tasks.}
    \label{fig:ft_res_1lead}
\end{figure}

\begin{figure}[h]
    \centering
    \includegraphics[width=\textwidth]{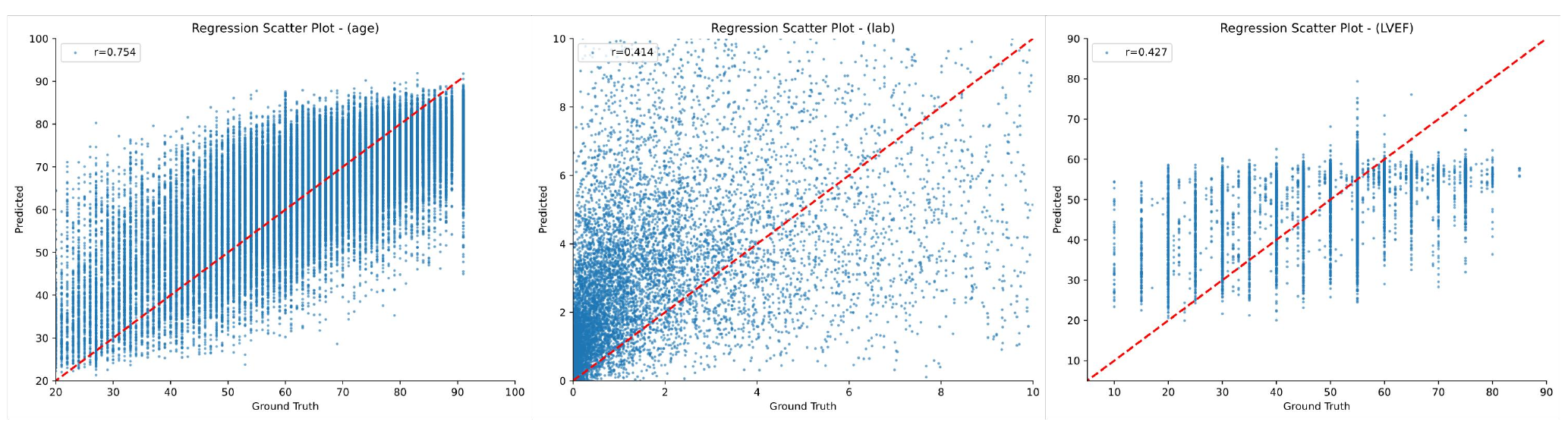} 
    \caption{Regression scatter of single-lead ECG downstream tasks.}
    \label{fig:ft_reg}
\end{figure}

Moreover, as shown in the figure \ref{fig:ft_res_1lead} and \ref{fig:ft_reg}, we evaluated several downstream tasks on single-lead ECGFounder thorough fine-tuning. 

Also, as shown in the fig S6, we compare the performance of linear probing versus full fine-tuning, particularly focusing on the downstream classification tasks of CHD and LVEF. The detailed results clearly demonstrate that full fine-tuning consistently achieves better performance compared to linear probing, particularly for the ECGFounder model. Furthermore, the ECGFounder model shows superior performance compared to SIMCLR and random initialization under both fine-tuning methods, underscoring the strong representational capacity of ECGFounder.

\begin{figure}[h]
    \centering
    \begin{subfigure}[b]{0.45\textwidth}
        \centering
        \includegraphics[width=\textwidth]{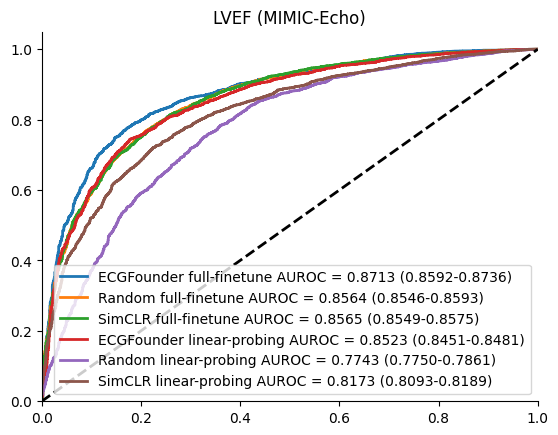}
        \label{fig:subfig1}
    \end{subfigure}
    \hfill
    \begin{subfigure}[b]{0.45\textwidth}
        \centering
        \includegraphics[width=\textwidth]{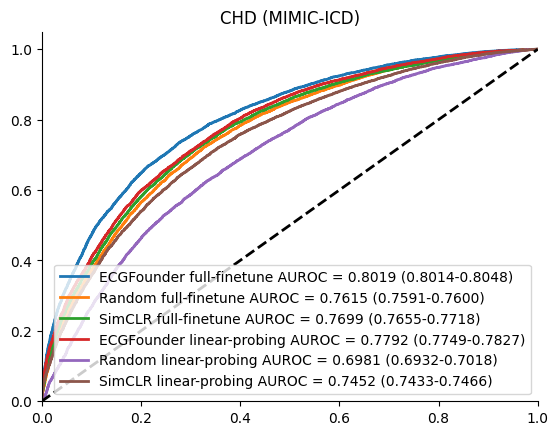}
        \label{fig:subfig2}
    \end{subfigure}
    \label{fig:diff_ft}
    \caption{Comparison of different fine-tuning methods, taking the classification of CHD and LVEF as examples.}
\end{figure}

\begin{table}[]

\centering
\caption{The PPV and prevalence of the fine-tuning task.}
\label{tab:PPV}
\begin{tabular}{l|cc}
\toprule
Task & PPV(\%)  & Prevalence(\%)  \\
\midrule
Age  & 77.1 & 53.9       \\
CHD  & 65   & 41.4       \\
CKD  & 24.2 & 13.2       \\
Lab  & 94.1 & 82.4       \\
LVEF & 94.1 & 76.9       \\
Sex  & 83.7 & 51.7      \\
\bottomrule
\end{tabular}
\end{table}

\begin{figure}[h]
    \centering
    \begin{subfigure}[b]{0.245\textwidth} 
        \includegraphics[width=\textwidth]{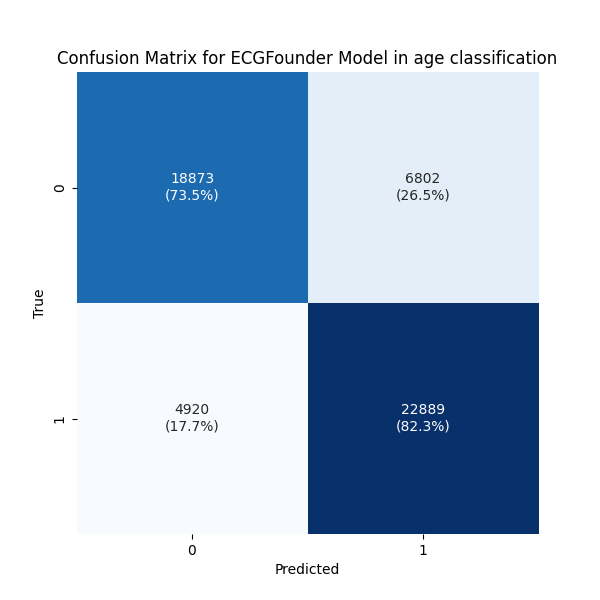} 
    \end{subfigure}
    \begin{subfigure}[b]{0.245\textwidth}
        \includegraphics[width=\textwidth]{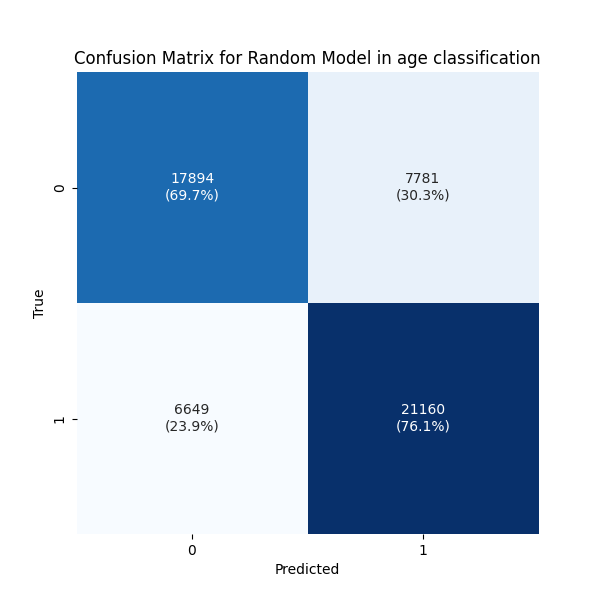}
    \end{subfigure}
    \begin{subfigure}[b]{0.245\textwidth}
        \includegraphics[width=\textwidth]{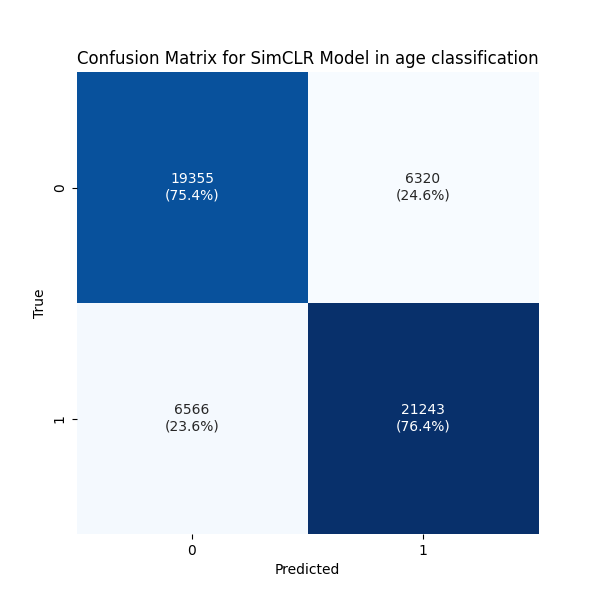}
    \end{subfigure}
    \begin{subfigure}[b]{0.245\textwidth}
        \includegraphics[width=\textwidth]{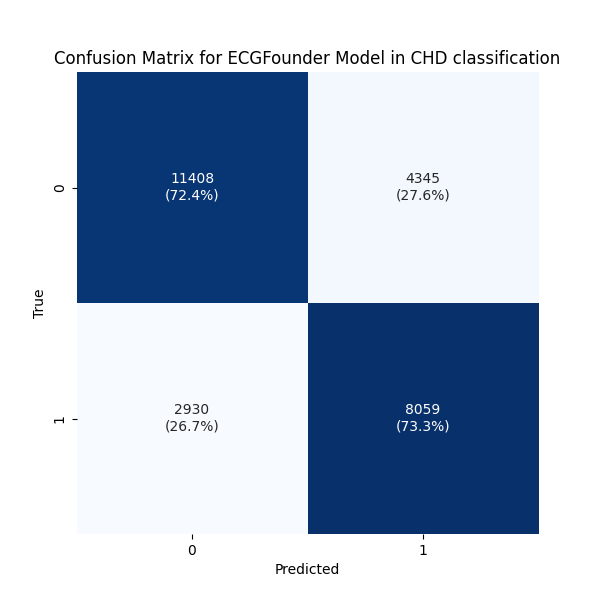}
    \end{subfigure}
    \begin{subfigure}[b]{0.245\textwidth}
        \includegraphics[width=\textwidth]{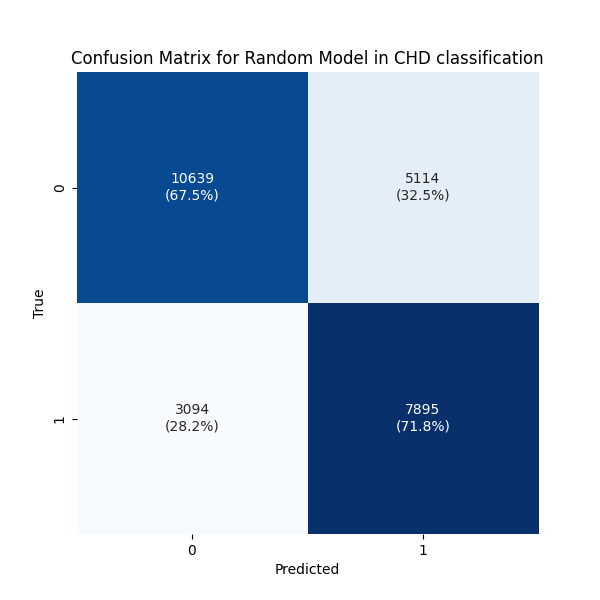}
    \end{subfigure}
    \begin{subfigure}[b]{0.245\textwidth}
        \includegraphics[width=\textwidth]{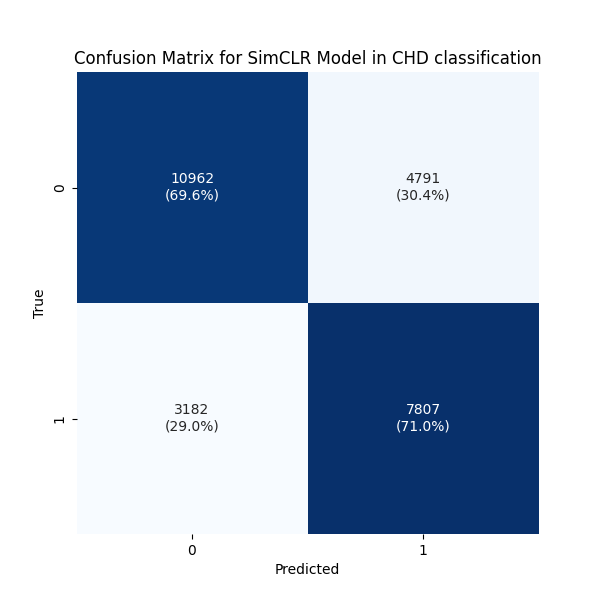}
    \end{subfigure}
    \begin{subfigure}[b]{0.245\textwidth}
        \includegraphics[width=\textwidth]{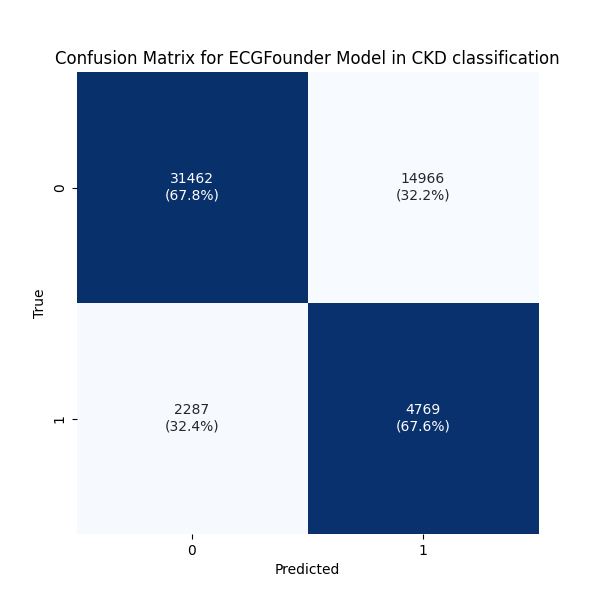}
    \end{subfigure}
    \begin{subfigure}[b]{0.245\textwidth}
        \includegraphics[width=\textwidth]{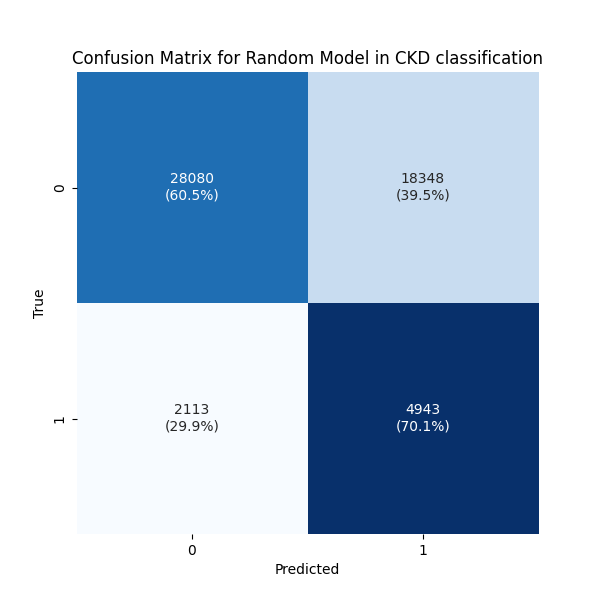}
    \end{subfigure}
    \begin{subfigure}[b]{0.245\textwidth}
        \includegraphics[width=\textwidth]{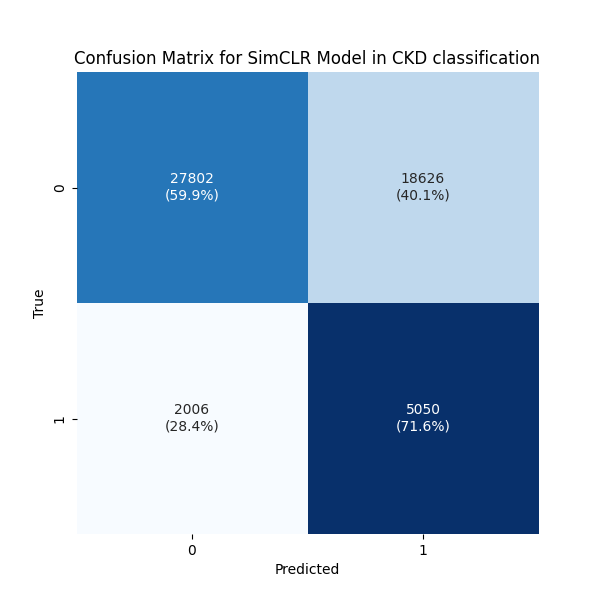}
    \end{subfigure}
    \begin{subfigure}[b]{0.245\textwidth}
        \includegraphics[width=\textwidth]{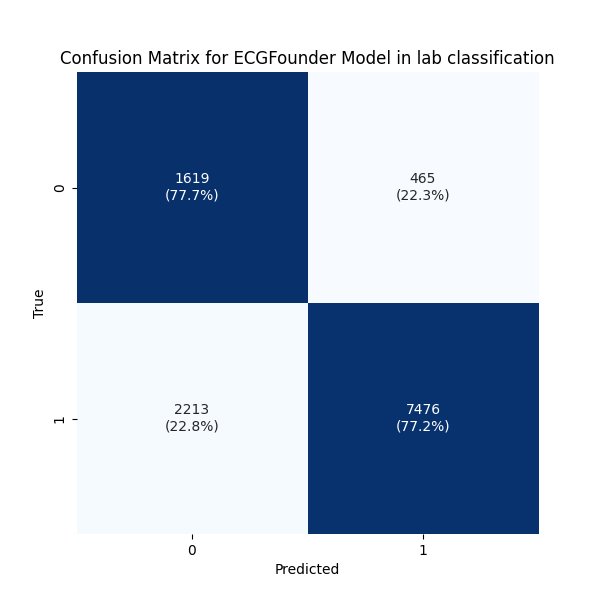}
    \end{subfigure}
    \begin{subfigure}[b]{0.245\textwidth}
        \includegraphics[width=\textwidth]{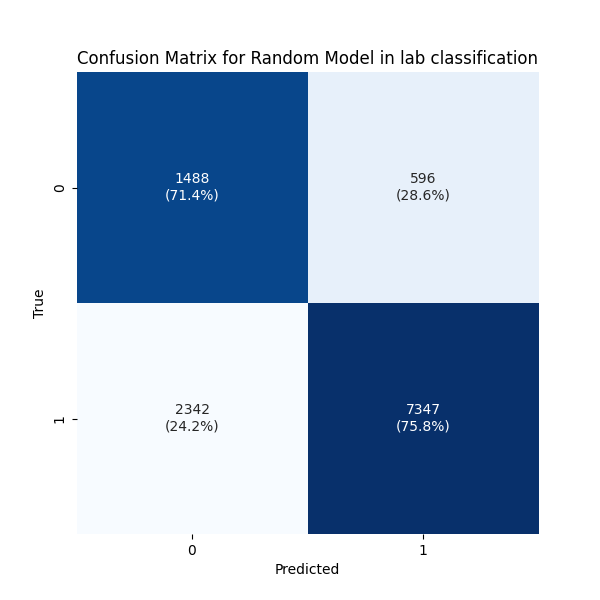}
    \end{subfigure}
    \begin{subfigure}[b]{0.245\textwidth}
        \includegraphics[width=\textwidth]{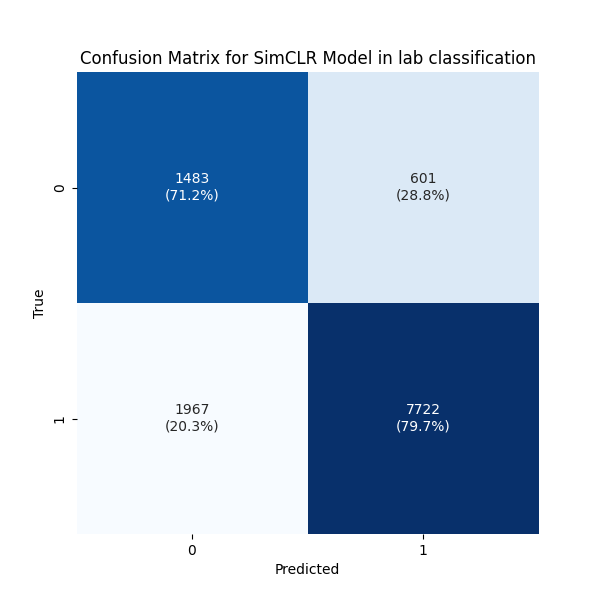}
    \end{subfigure}
    \begin{subfigure}[b]{0.245\textwidth}
        \includegraphics[width=\textwidth]{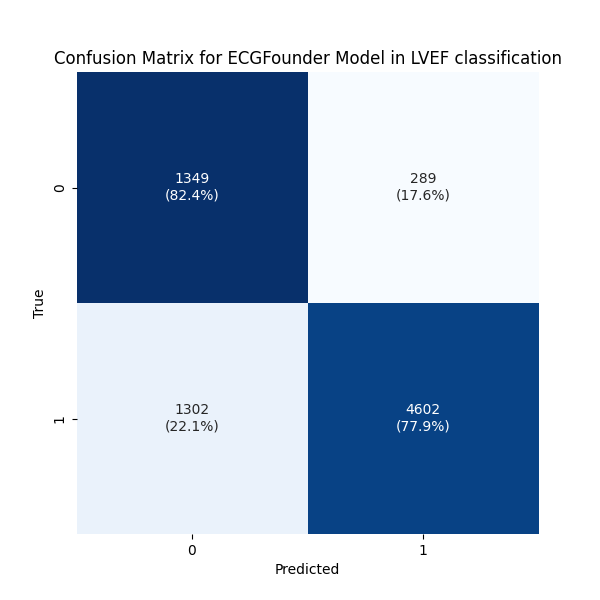}
    \end{subfigure}
    \begin{subfigure}[b]{0.245\textwidth}
        \includegraphics[width=\textwidth]{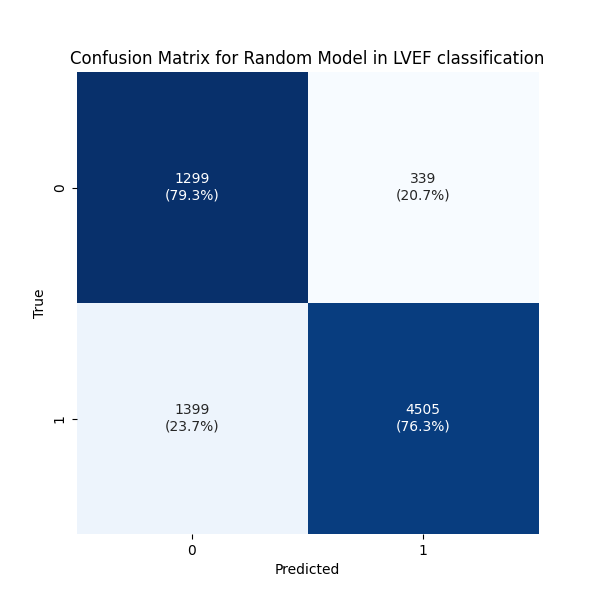}
    \end{subfigure}
    \begin{subfigure}[b]{0.245\textwidth}
        \includegraphics[width=\textwidth]{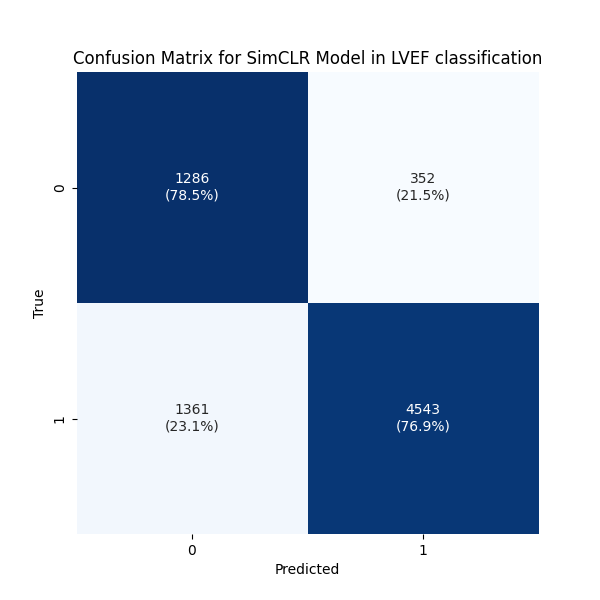}
    \end{subfigure}
    \begin{subfigure}[b]{0.245\textwidth}
        \includegraphics[width=\textwidth]{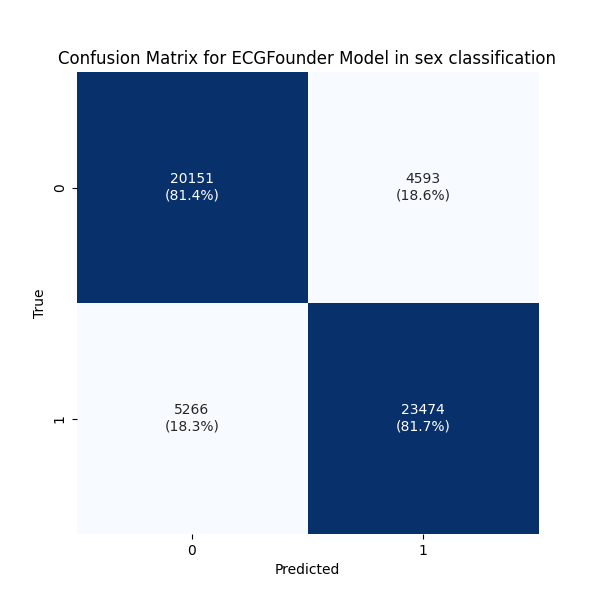}
    \end{subfigure}
    \begin{subfigure}[b]{0.245\textwidth}
        \includegraphics[width=\textwidth]{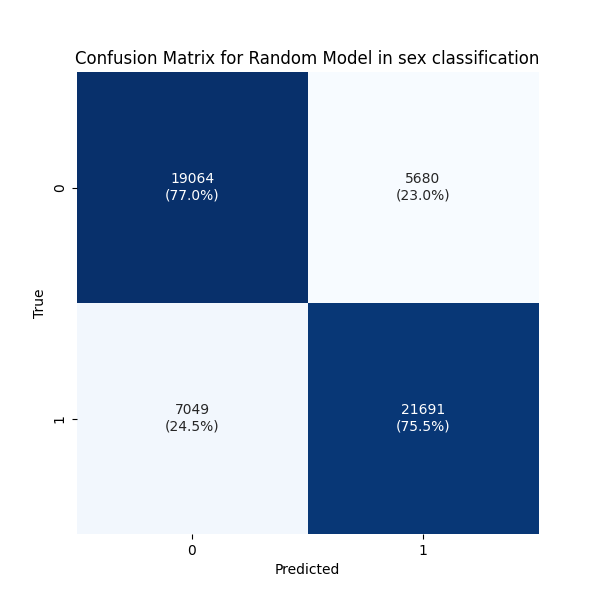}
    \end{subfigure}
    \begin{subfigure}[b]{0.245\textwidth}
        \includegraphics[width=\textwidth]{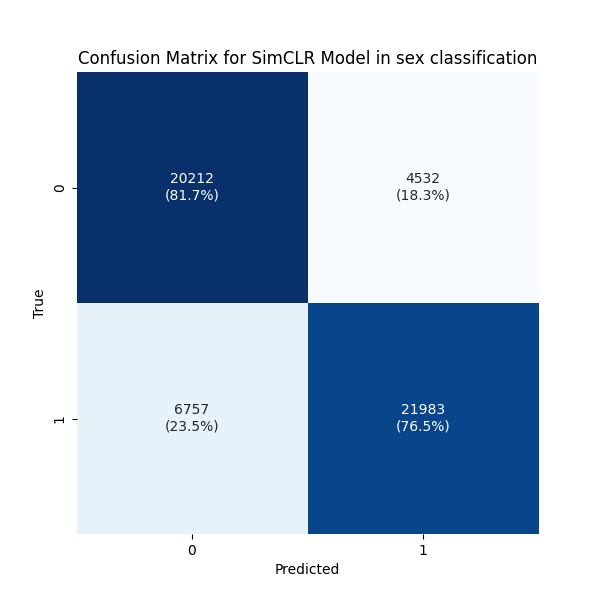}
    \end{subfigure}

    \caption{18 images arranged in a 3x6 grid.} 
\end{figure}

\newpage

\subsection*{S6. Data link}
\url{https://bdsp.io/content/heedb/1.0/}

\subsection*{S7. Model weights link}
 \url{https://huggingface.co/PKUDigitalHealth/ECGFounder}

\subsection*{S8. Code link}
\url{https://github.com/PKUDigitalHealth/ECGFounder}

\clearpage

\end{document}